\documentclass{article}

\usepackage[final,nonatbib]{hy}
\usepackage[numbers]{natbib}

\usepackage[utf8]{inputenc} 
\usepackage[T1]{fontenc}    
\usepackage{hyperref}  
\usepackage{url}        
\usepackage{booktabs}  
\usepackage{amsfonts}    
\usepackage{nicefrac}       
\usepackage{microtype}      
\usepackage{xcolor}   
\usepackage{tabularx}
\usepackage{graphicx}
\usepackage{xspace}
\usepackage{amsmath,amssymb,amsthm}
\usepackage{multirow}
\usepackage{colortbl}
\usepackage{pifont}         
\usepackage{arydshln}       
\usepackage{fontawesome5}
\usepackage{placeins}       

\usepackage[export]{adjustbox}
\usepackage[ruled]{algorithm2e}
\usepackage[inline, shortlabels]{enumitem}
\usepackage[T1]{fontenc}
\usepackage{hyperref}
\usepackage{microtype}
\usepackage{pifont}
\usepackage{xcolor}
\usepackage{xurl}
\usepackage{float}
\usepackage{graphicx}
\usepackage{booktabs}
\usepackage{tabularray}
\usepackage{makecell}
\usepackage{array}
\usepackage{rotating}
\usepackage{multicol}
\usepackage{multirow}
\usepackage{listings}
\usepackage{amsmath, amsfonts}
\usepackage{nicefrac}
\usepackage{subcaption}

\UseTblrLibrary{booktabs}

\makeatletter
\newcommand{\ssymbol}[1]{\@fnsymbol{#1}}
\newcommand{\romanNumeral}[1]{\expandafter\@slowromancap\romannumeral #1@}
\makeatother

\definecolor{linkblue}{HTML}{0055E9}
\hypersetup{colorlinks,urlcolor=linkblue,citecolor=linkblue,linkcolor=linkblue}

\makeatletter
\DeclareRobustCommand\onedot{\futurelet\@let@token\@onedot}
\def\@onedot{\ifx\@let@token.\else.\null\fi\xspace}

\makeatother

\def\shortname{\mbox{\textsc{Skill-Use}}\xspace}

\definecolor{citecolor}{rgb}{0.21,0.49,0.74}
\definecolor{linkcolor}{HTML}{ED1C24}
\definecolor{graycolor}{rgb}{0.95,0.95,0.95}

\usepackage[capitalize]{cleveref}
\crefname{section}{Sec.}{Secs.}
\crefname{table}{Tab.}{Tabs.}
\crefname{figure}{Fig.}{Figs.}

\newcommand{\xmark}{\ding{55}}
\newcommand{\cmark}{\ding{51}}

\theoremstyle{definition}
\newtheorem{definition}{Definition}

\newcommand{\ourdata}{\shortname}

\definecolor{findingsblue}{HTML}{EAF4FF}
\definecolor{findingsaccent}{HTML}{2878B5}
\newcommand{\findingsicon}{\textcolor{findingsaccent}{\faSearch}}
\newsavebox{\findingsboxsave}
\newenvironment{findingsbox}{%
  \par\smallskip\noindent
  \begin{lrbox}{\findingsboxsave}%
    \begin{minipage}{\dimexpr\linewidth-2\fboxsep\relax}%
      \findingsicon\hspace{0.35em}\textcolor{findingsaccent}{\textbf{Findings:}}\hspace{0.35em}\bfseries
}{%
    \end{minipage}%
  \end{lrbox}%
  \colorbox{findingsblue}{\usebox{\findingsboxsave}}%
  \par\smallskip
}

\renewcommand{\ghlink}{https://github.com/JinyiHan99/Skill-Use-Bench}

\makeatletter
\def\@maketitle{\vbox{\hsize\textwidth
 {\centering {\Large\bf \@title\par}}
\vspace{-0.1cm}
 \begin{center}\rule{\z@}{6pt}\@author\end{center}%
\vskip 0.15in minus 0.1in
}%
\begin{center}
\vspace{-0.6cm}
\begin{tabular}{rl}
\github & \url{\ghlink}\\
\end{tabular}
\end{center}
\thispagestyle{firstpage}}
\makeatother

\title{%
  \raisebox{-0.30\height}{\includegraphics[height=2.2em]{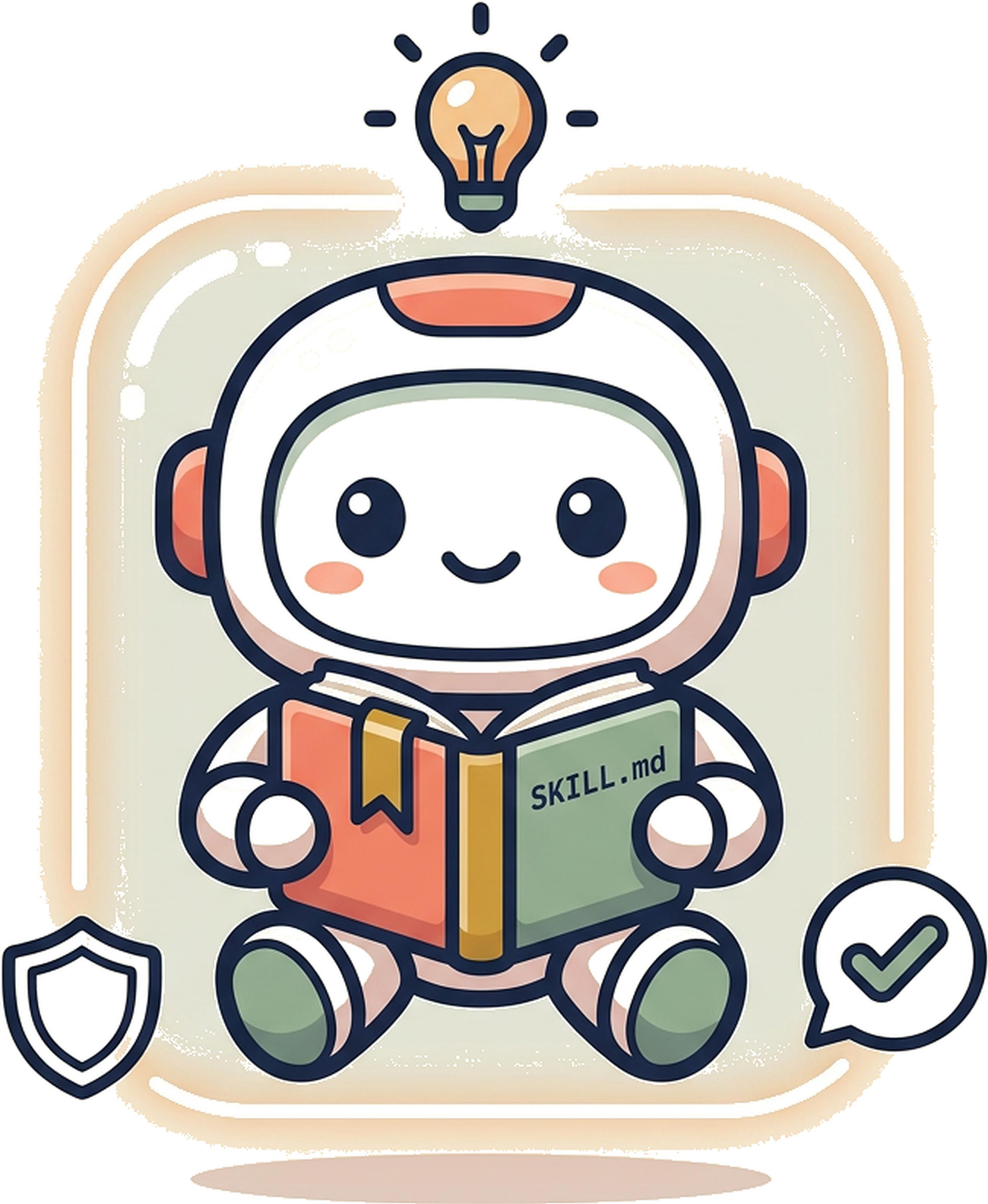}}\hspace{0.35em}%
  \shortname: Can LLMs Actually Use Skills in Agentic Harnesses?%
}

\author{%
  Jinyi Han\textsuperscript{1,\,$\ast$} \quad
  Yuanjian Xu\textsuperscript{2,\,$\ast$} \quad
  Ying Liao\textsuperscript{3} \quad
  Xinyi Wang\textsuperscript{3} \quad
  Zishang Jiang\textsuperscript{3} \\[0.35em]
  Zixiang Di\textsuperscript{1} \quad
  Fanyang Lu\textsuperscript{4} \quad
  Zhichao Hu\textsuperscript{4,\,$\dagger$} \quad
  Yanghua Xiao\textsuperscript{3,\,$\dagger$} \\[0.75em]
  {\normalfont\small
    \textsuperscript{1}East China Normal University \quad
    \textsuperscript{2}Hong Kong University of Science and Technology \\[0.15em]
    \textsuperscript{3}Fudan University \quad
    \textsuperscript{4}Tencent Hunyuan \\[0.35em]
    \textsuperscript{$\ast$}Equal contribution.\quad
    \textsuperscript{$\dagger$}Corresponding authors.
  }%
}

\begin{document}

\maketitle

\makeatletter
\def\@makefnmark{}
\makeatother

\begin{abstract}
Large language model (LLM) agents increasingly rely on skills, structured documents that specify when to act, which procedure to follow, and which tools are allowed. Existing evaluations mostly judge the quality of a skill or its contribution to task success, leaving unexamined whether an agent can recognize a relevant skill and apply it on its own. We introduce \ourdata, a benchmark that evaluates skill use under progressive disclosure, where an agent sees only a skill's name and short description and must retrieve the full procedure before following it. \ourdata separates three facets of skill use. Trigger measures whether the agent invokes the relevant skill, Compliance measures how faithfully it follows the prescribed procedure, and Boundary measures whether it avoids forbidden operations. A Skill-Use (SU) score combines the three and credits execution only after the skill is triggered. \ourdata pairs $79$ real skills with $177$ executable tasks across nine domains, each grounded in real files, run in an isolated Docker sandbox, and scored by a trajectory-based rubric. Evaluating eight LLMs under two agent harnesses, we find that reliable skill use remains out of reach, as the strongest configuration reaches an SU of only $0.613$. Triggering and procedural compliance fail as independent bottlenecks, and both scores and model rankings shift with the harness, so skill use behaves as a capability conditioned on the harness rather than a fixed property of the model.
\end{abstract}

\begin{figure*}[h]
    \centering
    \includegraphics[width=0.98\linewidth]{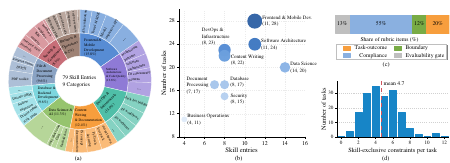}
    \caption{Composition and coverage of \ourdata. (a) Domain distribution of the $79$ skills across nine categories, where the inner ring gives each category's share and the outer ring lists representative skills. (b) Skill and task counts per domain, where each bubble marks one domain and reports its skill and task totals. (c) Composition of the rubric items by type. (d) Distribution of skill-exclusive requirements per task, our proxy for difficulty, where the dashed line marks the mean.}
    \label{fig:domain_dist}
\end{figure*}

\section{Introduction}
Large language model (LLM) agents increasingly carry out complex, multi-step tasks in real software and web environments~\citep{luo2025large,xu2024theagentcompany}. Within the past year, skills have become an emerging interface for agent harnesses, popularized by Claude Code~\citep{anthropic_claude_code_skills2025} and now appearing across systems such as Cursor, OpenClaw, and OpenAI Codex. Community repositories now host a fast-growing library of skills for diverse tasks~\citep{ling2026agent,li2026skillsbench,zhong2026skilllearnbench,li2026agentskillos}. These systems and repositories treat a skill as a structured document that encodes reusable domain knowledge for accomplishing a class of tasks, specifying the situations in which it applies, the procedure to follow, and the tools and restrictions that govern its execution. The implicit assumption is that pairing a capable agent with the right skill is all it takes to solve the corresponding task.

The growing skill ecosystem translates into real capability only if agents can use the skills they are given. In the progressive-disclosure mechanism~\citep{anthropic_claude_code_skills2025}, the agent sees only a skill's name and a short description; the full procedure is loaded only when the agent asks for it. To solve a task, the agent must first judge that a skill applies, retrieve it, and then follow it faithfully. 

Existing benchmarks, however, do not evaluate all of these requirements together. Two lines of work come closest, but each covers only part of the process. Skill-centric benchmarks treat skills as artifacts and judge their intrinsic quality or their marginal contribution to task success~\citep{li2026skillsbench,zhong2026skilllearnbench,liang2026skillnet,yang2026autoskill,zhou2026memento}; their evidence is the final outcome rather than the consuming agent's behavior. The closest effort scores adherence to the preconditions and orderings prescribed inside a skill, yet preloads the full skill and thereby bypasses retrieval altogether~\citep{chen2026slbench}. Instruction-following benchmarks score constraint compliance, but the constraints they target concern surface properties of the output such as format, length, or wording~\citep{zhou2023instruction,jiang2023followbench,qin2024infobench,wen2024complexbench,qin2025agentif}, and reach the agent verbatim in the prompt. A skill, in contrast, prescribes a domain procedure together with forbidden operations, and reaches the agent only through progressive disclosure. Neither line, therefore, addresses skill use as a process the agent must carry out on its own.

To fill this gap, we introduce \textbf{\ourdata}, a benchmark that evaluates skill use along three facets. {Trigger} measures whether an agent retrieves the relevant skill from its name and short description alone. {Compliance} measures adherence to the prescribed procedure, while {Boundary} measures avoidance of forbidden operations. By separating recognition, execution, and restraint, the benchmark reveals where skill use fails.
Specifically, we construct \ourdata in three stages. We curate real skills and design executable tasks whose procedures leave observable traces and whose forbidden shortcuts are genuinely tempting. We then author trajectory-based rubrics for each task, and finally validate every skill-task-rubric triple end to end. Masked-skill screening removes generic requirements, and multi-agent adversarial review audits task scope and rubric verifiability.

\ourdata pairs $79$ real skills from community repositories with $177$ executable tasks across nine domains (Figure~\ref{fig:domain_dist}). Each task is grounded in real files and runs in an isolated Docker sandbox with full tool access; its complete trajectory is logged and scored using a multi-item rubric. We report the three facets separately and combine them into a gated Skill-Use (SU) score, where execution receives credit only after the relevant skill is triggered.

Evaluating eight state-of-the-art models under two agent harnesses, we find that reliable skill use remains out of reach. The strongest configuration reaches an SU of only $0.613$. Triggering and procedural compliance break down as separate bottlenecks, since a model that recognizes a skill often still departs from its procedure while a model that could comply sometimes fails to trigger at all. Both absolute scores and model rankings shift once the harness changes, so skill use behaves as a capability conditioned on the surrounding harness rather than as a fixed property of the model~\citep{zhang2026harness,yao2026harnessbench}. 

\medskip
\noindent\textbf{Contributions.}
In summary, we make the following contributions:

\begin{itemize}
    \item We first formulate skill use as a distinct evaluation target and decompose it into three aspects, Trigger, Compliance, and Boundary, that capture recognition, procedural adherence, and restraint under progressive disclosure.
    \item We release \ourdata, a broad-coverage benchmark spanning nine domains, with $79$ real-world skills and $177$ sandboxed tasks graded by trajectory-based rubrics, together with a reusable three-stage construction pipeline.
    \item We conduct a systematic evaluation of eight frontier LLMs under two agent harnesses, yielding empirical insights into skill-use failure modes.
\end{itemize}

\section{Related Work}
\label{sec:related}

Our work is most closely related to two lines of research: skill benchmarks and instruction-following benchmarks. Table~\ref{tab:bench_compare} in Appendix~\ref{app:benchmark_details} provides a detailed comparison.

\paragraph{Skill Benchmarks.}
Skills package reusable agent knowledge as progressive-disclosure documents~\citep{anthropic_claude_code_skills2025,li2026skillsbench,li2026agentskillos}. SkillsBench~\citep{li2026skillsbench} evaluates whether a provided skill improves task completion. SkillLearnBench~\citep{zhong2026skilllearnbench} evaluates whether agents can generate reusable skills from experience through continual learning. Related work on skill libraries and continual skill acquisition focuses on organizing, producing, and refining such artifacts~\citep{liang2026skillnet, yang2026autoskill, zhou2026memento}. Collectively, these benchmarks focus on skill quality or final task success rather than whether an agent correctly uses a fixed skill. SLBench~\citep{chen2026slbench} is closest to our setting. It evaluates whether an agent respects logical relations within a skill, including preconditions, constraints, and overrides. However, preloading the full skill document leaves skill recognition and retrieval untested.

\paragraph{Instruction-Following Benchmarks.}
Instruction following measures whether LLMs can complete a task while satisfying user or system requirements. IFEval~\citep{zhou2023instruction} formalizes this ability as compliance with verifiable constraints on length, format, keywords, and response structure. Later benchmarks broaden the scope to richer constraint types~\citep{jiang2023followbench, qin2024infobench, wen2024complexbench}, system prompts~\citep{qin2024sysbench}, and multi-turn interactions~\citep{he2024multi}. AgentIF~\citep{qin2025agentif} further studies agentic instructions drawn from real system prompts. These benchmarks provide requirements directly in the prompt and therefore do not evaluate instruction following in the skill-use setting.

\paragraph{Agent Harness.}
Agent behavior depends not only on the base model but also on the surrounding harness that manages context, tools, and control flow. Recent studies show that the harness can account for more performance variance than the model and even reverse model rankings~\citep{zhang2026harness,yao2026harnessbench}. Motivated by this, we evaluate skill use across multiple harnesses rather than treating it as a fixed property of the model.

In summary, \ourdata extends instruction-following evaluation to progressively disclosed skills. It also moves skill benchmarking beyond artifact quality and task success toward process-level evaluation in an open environment.

\section{\ourdata Benchmark}
\label{sec:benchmark}

We begin by formalizing the benchmark setting and evaluation protocol in Section~\ref{sec:evaluation_protocol}. Section~\ref{sec:benchmark_coverage} then characterizes the benchmark's composition and coverage. Finally, Section~\ref{sec:construction} details the three-stage construction pipeline.

\subsection{Problem Formulation}
\label{sec:evaluation_protocol}
Each evaluation instance asks whether an agent can recognize and correctly apply one relevant skill while solving a realistic task. An instance is a tuple $x=(s,q,\mathcal{E},\mathcal{R})$ containing a skill $s$, a user task $q$, an executable sandbox $\mathcal{E}$, and a scoring rubric $\mathcal{R}$. The task consists of a user request and input assets such as a code repository, document, or dataset. The sandbox supplies the files and tools needed to execute the task. The rubric is hidden from the agent and is used only for evaluation.

\textbf{Input.}
We represent the skill as $s=(m,p)$, where $m$ contains its name, short description, and file path, and $p$ contains its full procedure. The agent initially receives $(q,\mathcal{E},m)$ as input. It does not receive $p$. Under the same progressive-disclosure interface~\citep{anthropic_claude_code_skills2025}, the agent must infer from the task and metadata that the skill is relevant, then retrieve the skill file to access its full procedure.

\textbf{Output.}
Running the agent on this input produces an execution trajectory
\begin{equation}
\tau=(o_0,a_1,o_1,\ldots,a_T,o_T)
\end{equation}
and a terminal artifact state $z_T$. Here, $o_0$ is the initial context, $a_k$ is the $k$-th action, and $o_k$ is the observation returned after that action. Actions include reading files, invoking tools, and editing artifacts. The state $z_T$ contains the files produced or modified by the agent when execution ends. We denote the recorded output by $y=(\tau,z_T)$.

\textbf{Scoring.}
Skill use unfolds in two stages: the agent must first recognize and retrieve the relevant skill, then apply its instructions faithfully. Trigger, formalized in Definition~\ref{def:trigger}, captures the first stage. Compliance, defined in Definition~\ref{def:compliance}, measures adherence to prescribed actions, while Boundary, defined in Definition~\ref{def:boundary}, measures respect for prohibited actions. Together, the three dimensions characterize whether a skill is engaged and how faithfully it governs the subsequent execution.

\begin{definition}[Trigger]
\label{def:trigger}
For output $y=(\tau,z_T)$ and target skill $s$, Trigger is $g(y;s):=\mathbb{I}[\operatorname{Retrieve}(s)\in\tau]$.
\end{definition}

\begin{definition}[Compliance]
\label{def:compliance}
Let $\mathcal{R}^{c}$ be the set of prescribed skill requirements. Each item $i$ has weight $w_i>0$ and verifier $\phi_i(y)=1$ iff the requirement is satisfied. Compliance is $\mathrm{C}(y):=\frac{\sum_{i\in\mathcal{R}^{c}}w_i\phi_i(y)}{\sum_{i\in\mathcal{R}^{c}}w_i}$.
\end{definition}

\begin{definition}[Boundary]
\label{def:boundary}
Let $\mathcal{R}^{b}$ be the set of skill prohibitions. Here $\phi_i(y)=1$ iff the prohibited behavior is absent. Boundary is $\mathrm{B}(y):=\frac{\sum_{i\in\mathcal{R}^{b}}w_i\phi_i(y)}{\sum_{i\in\mathcal{R}^{b}}w_i}$.
\end{definition}

Finally, we consolidate the three dimensions into a single score that summarizes overall skill-use quality:
\begin{equation}
\mathrm{SU}(y)=g(y;s)\left[
\alpha\,\mathrm{C}(y)
+(1-\alpha)\,\mathrm{B}(y)\right],
\label{eq:su}
\end{equation}
We fix $\alpha=0.7$ throughout the paper, giving Compliance the larger weight because it measures whether the agent enacts the skill's prescribed procedure, while Boundary only guards against prohibited actions. Sensitivity to this choice is reported in Appendix~\ref{app:alpha_sensitivity}.

\subsection{Benchmark Composition and Coverage}
\label{sec:benchmark_coverage}
\textbf{Domain Coverage.} \ourdata contains $79$ real skill entries paired with $177$ executable tasks across nine domains, spanning software development, infrastructure, data science, databases, document and content processing, security, and business operations. As shown in Figure~\ref{fig:domain_dist} (a) and (b), both skills and tasks are broadly distributed across these domains, with no single category dominating the benchmark.

\textbf{Rubric Composition.} Each task is graded by a hidden trajectory-level rubric. The $177$ rubrics hold $1{,}314$ scoring items, averaging $7.4$ per rubric and ranging from $3$ to $16$. Figure~\ref{fig:domain_dist}(c) groups these items into four types. Evaluability gates carry zero weight and only flag whether a run is well-formed enough to score. Compliance and Boundary are skill-derived and together form the Skill-Use signal. Task-outcome items record whether the requested result was produced and remain outside Skill-Use.

\textbf{Task Difficulty.} We proxy difficulty by the number of skill-exclusive requirements per task, those that follow from the skill itself rather than from the task prompt or generic practice. As Figure~\ref{fig:domain_dist}(d) shows, each task carries $4.8$ such requirements on average, with a tail reaching $12$. Most tasks therefore expose several independent failure points rather than a single checkpoint.

\subsection{Data Construction}
\label{sec:construction}

\begin{figure}[h]
\centering
\includegraphics[width=0.95\textwidth]{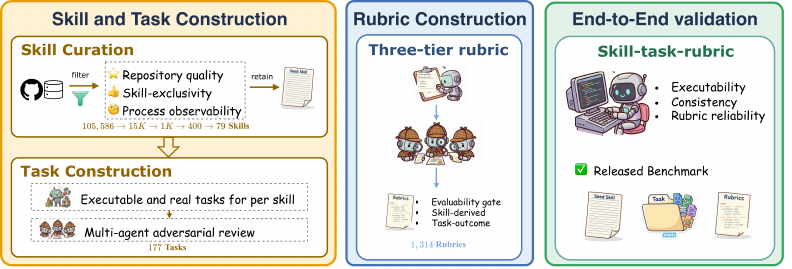}
\caption{Data construction pipeline of \ourdata. The stages run left to right. Skill and Task Construction curates $79$ skills from $105{,}586$ candidates and builds one in-scope task per skill, giving $177$ tasks. Rubric Construction attaches a three-tier rubric whose skill-derived items form the Skill-Use signal. End-to-End Validation checks each skill-task-rubric triple before release.}
\label{fig:construction}
\end{figure}

\ourdata is constructed in three stages. We first collect and curate real skill documents and construct executable tasks for the retained skills. We then design a trajectory-based scoring rubric for each task. Finally, we validate every skill-task-rubric triple end to end. Figure~\ref{fig:construction} gives an overview of the pipeline.

Skill curation, task construction, and rubric construction each follow a written guideline that we refine iteratively. We start from a short draft, apply it, audit the artifacts, and fold each uncovered failure back as a new rule, repeating until audits surface no new failure. The multi-agent adversarial review enforces the task and rubric guidelines. Appendix~\ref{app:guideline_dev} reports the protocol and the failures behind each rule.

\textbf{Skill and Task Construction.}
\label{sec:skill_curation}
Candidate skills are collected from public GitHub repositories and two existing benchmarks, SkillsBench~\citep{li2026skillsbench} and SkillLearnBench~\citep{ling2026agent}. We use real documents to preserve the structures, workflows, and tool assumptions encountered in practice, and retain skills according to two criteria:
\begin{itemize}
\item \textbf{Skill exclusivity.} The skill must contain requirements that cannot be inferred from the task request or generic practice, allowing the scored behavior to be attributed to the skill itself.
\item \textbf{Process observability.} The skill must constrain how the agent acts and leave verifiable evidence in its trajectory or artifacts, such as tool choices, action ordering, or prohibited operations.
\end{itemize}
We rank and sample candidates by repository quality and domain coverage, screen their process constraints, and apply a masked-skill check to remove generic requirements. Appendix~\ref{app:skill_pipeline} reports the funnel and its thresholds.
\label{sec:task_synthesis}
For each retained skill, we construct an executable task within its declared scope that exercises at least one skill-exclusive requirement, uses real assets, and hides the skill's prescribed procedure from the prompt. Two filters guard task quality. The masked-skill check drops any requirement a strong model still meets from the prompt alone, so every scored requirement remains skill-exclusive. A cross-family adversarial review then requires reviewers from different model families to confirm each such requirement, and we apply its revisions or discard tasks that cannot be repaired. Detailed descriptions are provided in Appendix~\ref{app:task_construction} and Appendix~\ref{app:adversarial_review}.

\textbf{Rubric Construction.}
\label{sec:rubric_synthesis}
For each task, we construct a weighted rubric with three tiers. Evaluability gates determine whether the run can be scored, skill-derived items measure Compliance and Boundary, and task-outcome items record whether the requested result was produced. Only skill-derived items contribute to Skill-Use. We design these items according to two criteria:
\begin{itemize}
\item \textbf{Skill exclusivity.} Every scored item must trace to a distinct skill requirement that is not implied by the task prompt or generic practice.
\item \textbf{Observable evidence.} Every item must be verified from a specific trajectory event or artifact. We prefer deterministic checks and use a scope-restricted LLM judge only when a requirement cannot be expressed reliably as a rule. Appendix~\ref{app:rubric_construction} specifies the rubric tiers and the judge.
\end{itemize}

Each rubric passes the same multi-agent adversarial review detailed in Appendix~\ref{app:adversarial_review}, and we apply its revisions before end-to-end validation.

\textbf{End-to-End Validation.}
\label{sec:e2e_validation}
Finally, we validate each assembled skill-task-rubric triple as a whole and retain only instances that satisfy three criteria:
\begin{itemize}
\item \textbf{Executability.} All required tools, dependencies, and assets are available in the isolated sandbox, allowing the task to run to completion under the evaluation harness.
\item \textbf{Cross-component consistency.} Each task requirement traces to a specific skill requirement, and each rubric item targets an observable signal that the task can produce during execution.
\item \textbf{Rubric reliability.} We replay each rubric on collected trajectories to identify inactive signals, unstable verifiers, format-sensitive failures, and leakage between Skill-Use and task-completion scoring.
\end{itemize}

\section{Experiments}
\label{sec:experiments}

\subsection{Experimental Setup}
\label{sec:setup}

\textbf{Models and harnesses.} We evaluate eight LLMs from seven families, namely Claude Opus 4.7, Claude Opus 4.8, GPT-5.5, DeepSeek-V4-Pro, GLM-5.1, MiniMax-M3, Kimi-K2.6, and Qwen3.6-Max. Each model is evaluated under two agent harnesses, Claude Code (CC) and Codex. Every task runs in an isolated Docker sandbox. Sampling parameters and gateway routing are reported in Appendix~\ref{app:setup}.

\textbf{Metrics.} We report Trigger, Compliance, Boundary, and their gated aggregate Skill-Use (SU), as defined in Section~\ref{sec:evaluation_protocol}. Model-level metrics are means over evaluation trajectories. Comp.$^{\dagger}$ and Bound.$^{\dagger}$ restrict to traces that trigger the target skill. Rubric items judged by an LLM use GPT-5.4 at temperature~$0$, following the standard LLM-as-a-judge practice~\citep{zheng2023judging}.

\newcommand{\championmark}{\raisebox{-0.25ex}{\includegraphics[height=1.65ex]{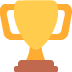}}}
\begin{table*}[h]
\centering
\footnotesize
\setlength{\tabcolsep}{4pt}
\renewcommand{\arraystretch}{1.0}
\resizebox{0.92\textwidth}{!}{%
\begin{tabular}{c l
  *{3}{w{c}{1.55cm}}
  >{\columncolor{findingsblue!60}}w{c}{1.55cm}
  *{2}{w{c}{1.55cm}}
  >{\columncolor{findingsblue!60}}w{c}{1.55cm}}
\toprule
\multirow{2}{*}{\textbf{Harness}} & \multirow{2}{*}{\textbf{Model}}
& \multicolumn{4}{c}{\textbf{Skill-Use}}
& \multicolumn{3}{c}{\textbf{On Triggered Traces}} \\
\cmidrule(lr){3-6}\cmidrule(lr){7-9}
& & Trigger & Compliance & Boundary & \textbf{SU} & Compliance$^{\dagger}$ & Boundary$^{\dagger}$ & \textbf{SU$^{+}$} \\
\midrule
\addlinespace[3pt]
\multirow{8}{*}{\rotatebox[origin=c]{90}{\textbf{Claude Code}}}
 & Claude Opus 4.7      & 0.966 & 0.609 & 0.712 & 0.599 & 0.606 & 0.709 & 0.637 \\
 & Claude Opus 4.8      & 0.940 & \textbf{0.625} & 0.704 & 0.597 & \textbf{0.638} & 0.710 & \textbf{0.660} \\
 & GPT-5.5\,\championmark & \textbf{0.972} & 0.611 & \textbf{0.718} & \textbf{0.613} & 0.625 & \textbf{0.723} & 0.654 \\
 & DeepSeek-V4-Pro      & 0.324 & 0.439 & 0.658 & 0.250 & 0.588 & 0.699 & 0.621 \\
 & GLM-5.1              & 0.324 & 0.455 & 0.639 & 0.267 & \textbf{0.638} & 0.632 & 0.636 \\
 & MiniMax-M3           & 0.833 & 0.569 & 0.706 & 0.501 & 0.592 & 0.698 & 0.624 \\
 & Kimi-K2.6            & 0.337 & 0.447 & 0.580 & 0.190 & 0.576 & 0.464 & 0.542 \\
 & Qwen3.6-Max          & 0.446 & 0.475 & 0.625 & 0.318 & 0.590 & 0.650 & 0.608 \\
\addlinespace[3pt]
\midrule
\addlinespace[3pt]
\multirow{8}{*}{\rotatebox[origin=c]{90}{\textbf{Codex}}}
 & Claude Opus 4.7      & 0.836 & \textbf{0.604} & 0.705 & 0.535 & \textbf{0.626} & \textbf{0.724} & \textbf{0.655} \\
 & Claude Opus 4.8\,\championmark & 0.881 & 0.599 & \textbf{0.715} & \textbf{0.559} & 0.626 & 0.721 & 0.654 \\
 & GPT-5.5              & \textbf{0.966} & 0.486 & 0.700 & 0.503 & 0.490 & 0.709 & 0.556 \\
 & DeepSeek-V4-Pro      & 0.650 & 0.496 & 0.653 & 0.379 & 0.571 & 0.701 & 0.610 \\
 & GLM-5.1              & 0.706 & 0.525 & 0.617 & 0.421 & 0.599 & 0.616 & 0.604 \\
 & MiniMax-M3           & 0.864 & 0.475 & 0.632 & 0.446 & 0.497 & 0.655 & 0.544 \\
 & Kimi-K2.6            & 0.706 & 0.489 & 0.623 & 0.374 & 0.529 & 0.614 & 0.554 \\
 & Qwen3.6-Max          & 0.684 & 0.266 & 0.573 & 0.238 & 0.304 & 0.571 & 0.384 \\
\addlinespace[3pt]
\bottomrule
\end{tabular}}
\caption{\textbf{Main results on \ourdata across Claude Code and Codex.} All metrics are scaled to $[0,1]$. Compliance and Boundary are averaged over all traces; their $\dagger$ variants are restricted to traces that trigger the target skill. \championmark{} marks the highest-SU model under each harness.}
\label{tab:main_results}
\vspace{-8pt}
\end{table*}

\subsection{Main Results}
\label{sec:main_results}

Table~\ref{tab:main_results} reports SU together with its Trigger, Compliance, and Boundary components under both harnesses. We make two main observations:

\begin{findingsbox}
Reliable skill use remains out of reach because triggering and conditional execution remain distinct bottlenecks.
\end{findingsbox}

The strongest configuration, GPT-5.5 under CC, achieves an SU of only $0.613$. Missed triggers alone do not explain this gap. Even among triggered traces, the highest Compliance$^{\dagger}$ reaches only $0.638$. Boundary is higher than Compliance across all model-harness configurations, so models avoid forbidden actions more reliably than they complete the full prescribed workflow.
Under CC, a group of open-weight middle-tier models trigger the target skill on only about a third of tasks, yet their conditional compliance is on par with models whose SU is twice as high. For them the missing capability is recognizing when the skill applies, not carrying it out. Under Codex, trigger rates concentrate in a narrower band and conditional compliance becomes the dominant source of between-model variation.

\begin{figure}[t]
\centering
\includegraphics[width=0.6\columnwidth]{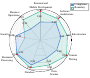}
\caption{Compliance and Boundary scores by skill category (CC). Boundary exceeds Compliance in every category except Security \& Compliance, where models follow the prescribed steps but fail to respect prohibitions.}
\label{fig:category_scores}
\vspace{-0.4cm}
\end{figure}

\begin{findingsbox}
Skill-use performance and rankings depend on the harness because CC and Codex stress different sub-abilities.
\end{findingsbox}

Absolute scores and relative orderings both shift when the harness changes. The head of the ranking is occupied by GPT-5.5 under CC and by Claude Opus 4.8 under Codex; several models improve under Codex while others regress, and the correlation between the two per-model SU vectors is only moderate, as detailed in Appendix~\ref{app:harness_delta}. SU is therefore a property of a model-harness configuration, and rankings under one harness may not transfer to another.
Moving from CC to Codex compresses the spread of Trigger while widening the spread of conditional Compliance. Codex lifts the weakest triggers toward the pack, yet some strong models lose ground on execution quality, with GPT-5.5's Compliance$^{\dagger}$ falling by more than a tenth. Codex thus lowers the barrier to selecting a skill but demands more sustained procedural execution once the skill is entered. Models whose strength is triggering benefit from CC, while models whose strength is execution retain more of their score under Codex.


\subsection{How Skill Type Affects Skill-Use}
\label{sec:category}

To examine how skill-use performance varies across skill categories, Figure~\ref{fig:category_scores} reports category-level Compliance and Boundary.

\begin{findingsbox}
Models follow concrete procedures more reliably than open-ended ones and generally avoid prohibitions more easily than they complete required workflows.
\end{findingsbox}

File processing, database, and data science skills have the highest Compliance. These skills typically prescribe a named tool, file format, or well-defined operation. Business analysis and software architecture score lower because their workflows involve open-ended planning. Boundary exceeds Compliance in almost every category, confirming that models suppress banned actions more easily than they execute every required step. Security and compliance is the sole exception. Models in this category achieve reasonable Compliance but the lowest Boundary, indicating that they carry out the safety steps yet still invoke the skill where restraint is needed.

\subsection{How Skill Injection Mode Affects Skill-Use}
\label{sec:mechanism}
\begin{figure}[t!]
\centering
\includegraphics[width=0.6\columnwidth]{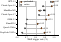}
\caption{Per-model Trigger rate under both injection modes on paired traces. Preloading consistently raises Trigger.}
\label{fig:mech_delivery_1}
\vspace{-0.3cm}
\end{figure}

\begin{figure}[t]
\centering
\includegraphics[width=0.6\columnwidth]{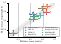}
\caption{Native SU ($x$) against preloaded SU ($y$). Per-model SU restricted to traces that trigger under both modes. Clustering near the diagonal indicates that injection mode has little effect on execution quality after the skill is triggered.}
\label{fig:mech_delivery_2}
\vspace{-0.4cm}
\end{figure}

To isolate the effect of skill injection, we pair two Codex runs on every (task, model) combination, changing only how the skill enters the agent's context. In the native mode, the prompt lists the skill's name, one-line description, and file path; the model must actively open the file to see the full procedure. In the preloaded mode, the entire skill document is inserted into the initial instructions so the procedure is visible from the start. Figures~\ref{fig:mech_delivery_1} and~\ref{fig:mech_delivery_2} report Trigger and SU under both modes.

\begin{findingsbox}
Preloading primarily improves skill selection, revealing retrieval as the main bottleneck under native injection.
\end{findingsbox}

On all paired traces, preloading improves SU because it raises Trigger. The effect is largest for models whose native Trigger is weak and nearly disappears for models already near saturation. When both modes trigger, the SU gap becomes small and most model points stay close to the diagonal. Thus, showing the full skill text mainly helps the model recognize the skill as relevant; it does not materially improve rule following once the skill has been selected.

Native injection asks the model to act from a compact skill description before it sees the full procedure. The paired comparison shows that this extra decision, rather than downstream execution, accounts for most of the injection effect.


\subsection{How Skill Library Size Affects Skill-Use}
\label{sec:scaling}

\begin{figure*}[t]
\centering
\includegraphics[width=1.0\textwidth]{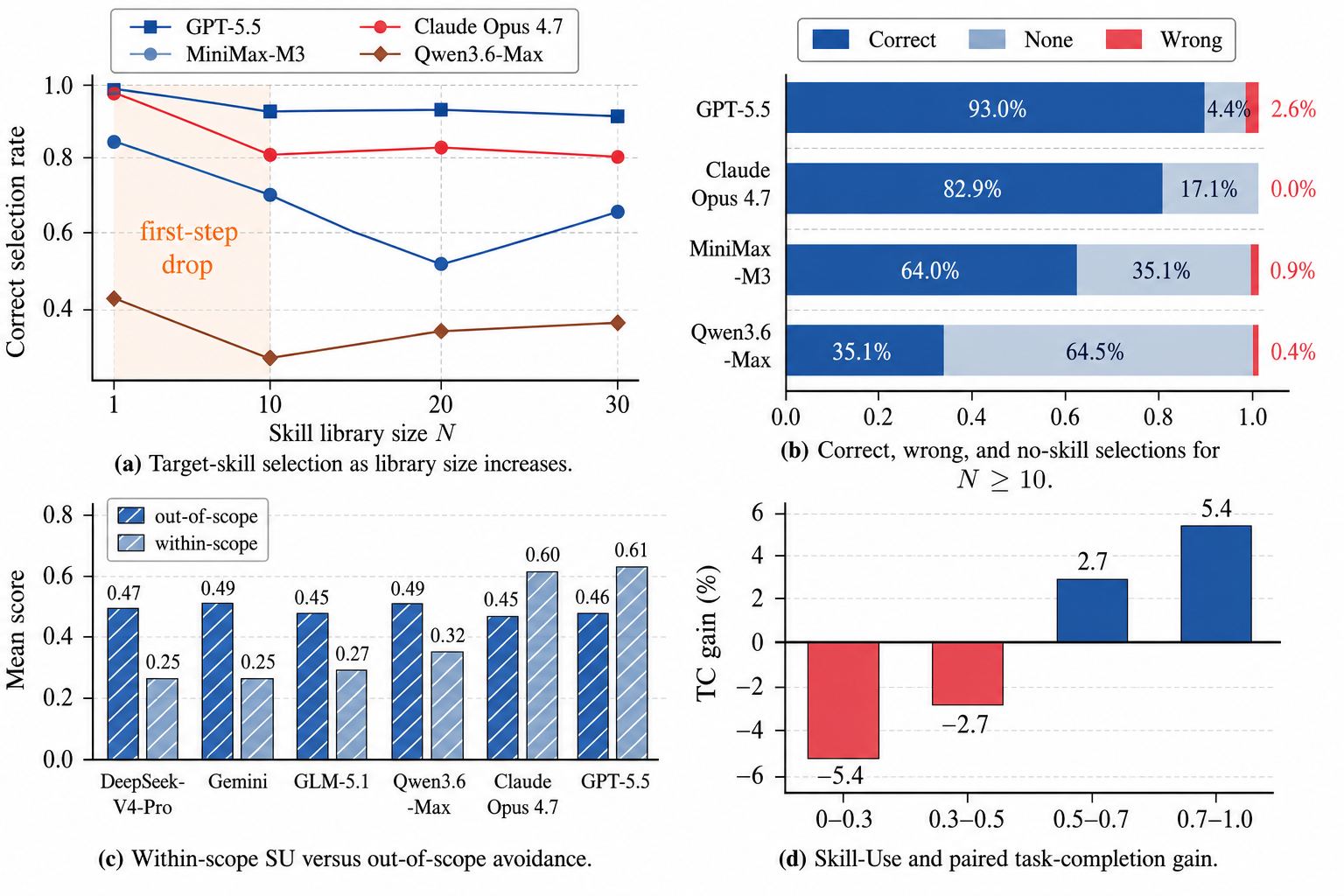}
\caption{Skill selection, applicability, and task-completion effects under CC. (a--b) Increasing library size mainly raises no-skill outcomes rather than wrong-skill selections. (c) Within-scope Skill-Use and out-of-scope avoidance are not aligned across models. (d) Paired task-completion gain increases with Skill-Use; each triggered skill-enabled trace is paired with a no-library baseline on the same task and model.}
\label{fig:scaling_boundary_tc}
\vspace{-0.4cm}
\end{figure*}

We test whether a larger skill library makes it harder to trigger the correct skill. Under CC, we install $N\!\in\!\{1, 10, 20, 30\}$ skills into the agent's library. The target skill is always included; the remaining $N{-}1$ slots are filled with randomly sampled distractors. We label a run correct if the model selects the target skill, wrong if it selects a distractor, and none if it does not invoke any skill. The results are shown in Figure~\ref{fig:scaling_boundary_tc}(a--b).

\begin{findingsbox}
Adding distractors causes the main performance drop, driven more by missed triggers than by wrong-skill selections.
\end{findingsbox}

All models mostly select the target reliably in the single-skill setting. Moving to a ten-skill library causes the visible drop, especially for weaker models. After that point, the curves change little. Library growth creates an entry cost, not a steady decline.
At larger library sizes, most failures are runs with no skill call. Wrong-skill calls are rare and occur mainly for broad, overlapping descriptions such as test-driven-development and software-architecture, as documented in Appendix~\ref{app:scaling_full}. Models often fail to decide that any skill is relevant enough to invoke.

\subsection{How Models Avoid Unnecessary Skill Use}
\label{sec:boundary}

All previous experiments use tasks that fall within the target skill's scope. Here we ask the question: can models refrain from invoking a skill when it does not apply? We construct $53$ out-of-scope tasks whose topics overlap a skill but whose requested outputs do not require that skill's procedure. We evaluate six models under CC and define an avoidance score as the fraction of out-of-scope tasks on which the model does not invoke the target skill. Figure~\ref{fig:scaling_boundary_tc}(c) compares each model's avoidance score with its within-scope SU.
\begin{findingsbox}
Strong skill use does not guarantee restraint when skills do not apply.
\end{findingsbox}

The rank correlation between within-scope SU and out-of-scope avoidance is negative, so the models best at following a skill can be the worst at suppressing unnecessary invocations. The two abilities rest on separate judgments. One governs behavior after entering a skill, and the other governs whether to enter at all. The over-invocations are driven by topical overlap, mostly document-shaped requests whose topic matches the skill's domain even though the output never needs its procedure, as detailed in Appendix~\ref{app:negative_full}. 

\subsection{How Skill-Use Relates to Task Completion}
\label{sec:su_tc_relation}

We test whether following a skill's procedure improves task completion, as shown in Figure~\ref{fig:scaling_boundary_tc}(d). Under CC, we start from the $779$ runs that trigger the target skill. For each run, we build a paired baseline on the same task and model with the skill library disabled. We group the pairs by SU and report the mean paired gain $\mathrm{TC}_{\mathrm{skill}}-\mathrm{TC}_{\mathrm{base}}$ in each bin.

\begin{findingsbox}
Whether a skill helps or hurts depends on both skill-use quality and skill type.
\end{findingsbox}

The paired gain turns from negative to positive near an SU score of $0.5$. Below this point, the model uses the skill partially or incorrectly, and task completion drops below the no-skill baseline. Above it, the model follows the procedure closely, and the skill acts as a scaffold that raises completion. A partial procedure is worse than none because the model commits to a prescribed toolchain or format without carrying it through, and the resulting artifact is neither what the skill demands nor what the model would produce unaided.
Adding a skill library is therefore not a free improvement; it pays off only when a model can execute the procedure and when the skill supplies an operational recipe rather than a stylistic prescription.


\FloatBarrier

\section{Conclusion}
\label{sec:conclusion}

This paper introduces \ourdata, a benchmark for evaluating skill use in LLM agents under progressive disclosure. \ourdata decomposes skill use into Trigger, Compliance, and Boundary, and pairs $79$ real skills with $177$ executable tasks across nine domains. Using this benchmark, we evaluate eight frontier LLMs under two agent harnesses. The results show that skill use remains an open problem: no configuration achieves reliable performance on all three axes, and both scores and rankings depend on the harness rather than on the model alone. Our analysis further indicates that the main gap is retrieval under progressive disclosure, and that within-scope skill use does not transfer to restraint on out-of-scope tasks. We release \ourdata, together with its construction pipeline, to support future work on training and evaluating agents that use skills reliably.

\clearpage

{\small
\bibliographystyle{unsrtnat}
\bibliography{ref}
}

\clearpage
\appendix

\setcounter{secnumdepth}{2}
\renewcommand{\thesection}{\Alph{section}}
\renewcommand{\thesubsection}{\thesection.\arabic{subsection}}

\newcommand{\tocA}[2]{%
  \par\smallskip
  \noindent\textcolor{findingsaccent}{\textbf{\ref{#1}}}\hspace{0.65em}%
  \textbf{#2}\nobreak\hspace{0.5em}\dotfill\hspace{0.5em}%
  \textcolor{findingsaccent}{\textbf{\pageref{#1}}}\par}
\newcommand{\tocB}[2]{%
  \noindent\hspace*{1.7em}%
  \textcolor{gray}{\ref{#1}}\hspace{0.6em}#2%
  \nobreak\hspace{0.5em}\dotfill\hspace{0.5em}%
  \textcolor{gray}{\pageref{#1}}\par}

\begin{center}
{\Large\bfseries Appendix Contents}
\end{center}
\vspace{1pt}
\vspace{3pt}

\tocA{app:benchmark_details}{Benchmark Comparison}
\tocA{app:pipeline}{Construction Pipeline}
\tocB{app:skill_pipeline}{Skill Selection}
\tocB{app:task_construction}{Task Construction}
\tocB{app:rubric_construction}{Rubric and Judge}
\tocB{app:adversarial_review}{Adversarial Review}
\tocB{app:e2e_validation}{Release Validation}
\tocB{app:guideline_dev}{Guideline Development}
\tocA{app:human_audit}{Human Quality Audit}
\tocA{app:setup}{Experimental Setup}
\tocA{app:analysis}{Additional Analysis}
\tocB{app:harness_delta}{Cross-Harness Behaviour}
\tocB{app:mechanism_full}{Skill Injection}
\tocB{app:scaling_full}{Library Scaling}
\tocB{app:negative_full}{Out-of-Scope Behaviour}
\tocB{app:robust_full}{Robustness Analysis}
\tocB{app:alpha_sensitivity}{Weight Sensitivity Analysis}
\tocA{app:prompts}{Prompt Templates}
\tocB{app:pr_skill_attacker}{Skill-Selection Red Team}
\tocB{app:pr_review}{Task and Rubric Adversarial Review}
\tocB{app:judge_prompt}{LLM Judge}
\tocA{app:ethics}{Ethical Statement}
\vspace{3pt}
\vspace{5pt}

\section{Benchmark Comparison}
\label{app:benchmark_details}

Table~\ref{tab:bench_compare} places \ourdata against instruction-following and skill-centric benchmarks. Instruction-following benchmarks put constraints in the prompt and grade the final response, so they neither run an agent in an open environment nor test recognition. Skill-centric benchmarks such as SkillsBench and SkillLearnBench focus on skill generation, skill learning, or task success under a supplied skill, rather than autonomous retrieval and process-level use. SLBench~\citep{chen2026slbench} checks procedure following with the full document preloaded, so recognition is assumed and Boundary is not isolated. \ourdata is the only benchmark that jointly scores Trigger, Compliance, and Boundary in an open, process-level setting under progressive disclosure. 


\providecommand{\cmark}{\ding{51}}
\providecommand{\xmark}{\ding{55}}
\begin{table*}[t]
    \centering
    \small
    \setlength{\tabcolsep}{5pt}
    \resizebox{\textwidth}{!}{%
    \begin{tabular}{l cc c ccc c}
    \toprule
    \multirow{2}{*}{\textbf{Benchmark}} & \multicolumn{2}{c}{\textbf{Environment}} & & \multicolumn{3}{c}{\textbf{Skill-Use Dimensions}} & \multirow{2}{*}{\textbf{Scale}} \\
    \cmidrule(lr){2-3}\cmidrule(lr){5-7}
    & \textbf{Open Env.} & \textbf{Process} & \textbf{Skill Doc.} & \textbf{Trigger} & \textbf{Compliance} & \textbf{Boundary} & \\
    \midrule
    IFEval~\citep{zhou2023instruction}          & \xmark & \xmark & \xmark & \xmark & \cmark & \xmark & $541$ inst. \\
    FollowBench~\citep{jiang2023followbench}    & \xmark & \xmark & \xmark & \xmark & \cmark & \xmark & $820$ inst. \\
    ComplexBench~\citep{wen2024complexbench}    & \xmark & \xmark & \xmark & \xmark & \cmark & \xmark & $1{,}150$ inst. \\
    InfoBench~\citep{qin2024infobench}          & \xmark & \xmark & \xmark & \xmark & \cmark & \xmark & $500$ inst. \\
    SysBench~\citep{qin2024sysbench}            & \xmark & \xmark & \xmark & \xmark & \cmark & \xmark & $500$ sys. \\
    AgentIF~\citep{qin2025agentif}              & \xmark & \xmark & \xmark & \xmark & \cmark & \xmark & $707$ inst. \\
    SOPBench~\citep{li2025agentorca}            & \xmark & \cmark & \xmark & \xmark & \cmark & \cmark & $663$ tasks \\
    \midrule
    SkillsBench~\citep{li2026skillsbench}       & \cmark & \xmark & \cmark & \xmark & \xmark & \xmark & $87$ tasks \\
    SkillLearnBench~\citep{zhong2026skilllearnbench} & \cmark & \xmark & \cmark & \xmark & \xmark & \xmark & $20$ tasks \\
    SLBench~\citep{chen2026slbench}             & \cmark & \cmark & \cmark & \xmark & \cmark & \cmark & $86$ tasks \\
    \midrule
    \ourdata (ours)                             & \cmark & \cmark & \cmark & \cmark & \cmark & \cmark & \textbf{$79$ skills / $177$ tasks / $1{,}314$ items} \\
    \bottomrule
    \end{tabular}
    }
    \caption{Positioning of \ourdata against instruction-following and skill benchmarks. \emph{Open Env.}: the agent acts in a sandbox with real tools. \emph{Process}: scoring uses the execution trajectory, not only the final answer. \emph{Skill Doc.}: guidance ships as a progressive-disclosure skill document rather than inline prompt text. \emph{Trigger}, \emph{Compliance}, and \emph{Boundary} are the three Skill-Use dimensions: recognizing and retrieving the skill, following its procedure, and respecting its prohibitions. \cmark~supported, \xmark~not supported. \ourdata is the only benchmark that covers all three dimensions in an open, process-level setting under progressive disclosure.}
    \label{tab:bench_compare}
\end{table*}

\section{Construction Pipeline}
\label{app:pipeline}

This section details each construction stage summarized in Section~\ref{sec:construction}, from skill selection through the guideline loop that governs it.

\subsection{Skill Selection}
\label{app:skill_pipeline}

Four phases narrow $105{,}586$ public GitHub skills to $79$. Cost per candidate rises with each phase, so cheap filters run first. Table~\ref{tab:skill_funnel} summarises the funnel.

\begin{table}[h]
\centering
\small
\renewcommand{\arraystretch}{1.2}
\begin{tabular}{@{}l r r l@{}}
\toprule
Phase & In & Out & Main rejection \\
\midrule
1. Filter + cluster  & $105{,}586$ & $15{,}000$  & low stars, duplicates \\
2. Fetch under quota & $1{,}500$   & $1{,}000$   & fetch failure \\
3. Rule scoring      & $1{,}000$   & $400$       & generic, no rule \\
4. Masked-skill eval & $400$       & $80$--$150$ & no exclusive rule \\
\bottomrule
\end{tabular}
\vspace{4pt}
\caption{\textbf{Skill selection funnel.} Input and output counts (approximate) and dominant rejection reason for each of the four phases that reduce $105{,}586$ candidate GitHub skills to $79$. Phase~1 output is downsampled to the Phase~2 fetch quota; the final set adds retained SkillsBench and SkillLearnBench entries.}
\label{tab:skill_funnel}
\end{table}

\textbf{Phase 1: filter and cluster.} We drop entries below $5$ stars, mirror aggregators, meta-skills that teach how to write skills, names that are pure numbers or single letters or contain \texttt{test}, \texttt{example}, \texttt{template}, or \texttt{demo}, and names a prior benchmark already covers. We impose no fixed taxonomy, since a wrong one would discard dense regions of $10^5$ unknown skills. Instead we split names on hyphens and underscores, cluster the top $200$ tokens, and ask an LLM to propose $15$ to $25$ categories grounded in those clusters. Keyword rules then label the pool, with the LLM as fallback.

\textbf{Phase 2: fetch under quotas.} Each category gets a fetch quota by its overlap with prior benchmarks: $80$ for no overlap, $50$ for partial, $20$ for existing coverage, capped near $1{,}500$. Within a category we rank by name specificity, then star count, then recency, so a compound name such as \texttt{postgres-migration-patterns} outranks a short generic one. For each entry we fetch \texttt{SKILL.md} first, then \texttt{.cursorrules} or \texttt{.clinerules}, then a constraint-style \texttt{README.md}.

\textbf{Phase 3: rule scoring.} We score each document $d$ on constraint density and style specificity,
\[
s_{\text{den}}(d) = \sum_{i} w_i^{\text{den}}\, f_i(d), \qquad
s_{\text{sty}}(d) = \sum_{j} w_j^{\text{sty}}\, g_j(d),
\]
and admit it only if it clears both thresholds, $s_{\text{den}}(d)\ge\tau_{\text{den}}$ and $s_{\text{sty}}(d)\ge\tau_{\text{sty}}$, with $\tau_{\text{den}}=0.30$ and $\tau_{\text{sty}}=0.25$. Density features $\{f_i\}$ count modal terms (\textsc{must}, \textsc{never}, \textsc{always}, \textsc{required}, \textsc{forbidden}, \textsc{critical}), code examples, good-versus-bad blocks, numbered rules, length, and guardrail sections. Style features $\{g_j\}$ count named tools and commands, process constraints, counter-intuitive rules, and machine-checkable cues. Thresholding the axes separately rejects documents that are dense but generic or specific but thin. We flag hard-to-sandbox needs (GUI, GPU, paid API) for Phase~4 without rejecting.

\textbf{Phase 4: masked-skill evaluation.} An LLM inspects every constraint of each surviving document, and the red-team attacker prompt in Figure~\ref{fig:prompt_skill_attacker} tries to refute each proposed constraint. Let $\mathcal{C}(d)$ be its constraint set, $\mathcal{C}_{\text{ex}}(d)$ those that survive the masked-skill check as non-generic, $\mathcal{C}_{\text{rv}}(d)$ those with a rule-based verifier (regex, file check, AST, command trace, or state diff), and $\mathcal{D}(d)$ the dimensions (tool choice, step ordering, conditional trigger, safety boundary, output format, error handling) they cover. We tier each document by
\[
\text{tier}(d) =
\begin{cases}
\text{A} & |\mathcal{C}_{\text{ex}}| \ge 3,\ |\mathcal{C}_{\text{rv}}| \ge 2,\ |\mathcal{D}| \ge 2, \\
\text{B} & |\mathcal{C}_{\text{ex}}| \ge 2,\ |\mathcal{C}_{\text{rv}}| \ge 1, \\
\text{C} & |\mathcal{C}_{\text{ex}}| \ge 1,\ |\mathcal{C}_{\text{rv}}| \ge 1, \\
\text{D} & \text{otherwise (rejected).}
\end{cases}
\]
We admit Tier~A without quota; Tiers~B and~C enter under the Phase~2 quotas, with looser caps for domains prior benchmarks do not cover. A $10\%$ human spot-check audits the LLM's labels.

\subsection{Task Construction}
\label{app:task_construction}

For each skill we write executable, in-scope tasks that a strong model cannot solve without the skill. The masked-skill check enforces this: we hide the document and ask whether a strong model, given only the prompt and generic practice, would produce the same behaviour. A requirement it still meets is generic and dropped; a requirement it fails is skill-exclusive and kept. We run the check twice, once per skill during selection and once per task-requirement pair during construction.

The check tests precision, not topic recognition. A good task turns on the exact choice the skill prescribes, such as tiering a rate limit per user rather than per IP at the skill's threshold, not on whether the model knows the general topic. Any requirement a model meets from common practice carries no signal.

Prompts read like user requests and hide the procedure. We reject checklist leakage that restates the rubric axes, directional leakage that reveals the intended fix, and reverse leakage that tells an out-of-scope task not to use the skill. We also keep prescribed formats, tool names, and parameters out of the prompt, since these are themselves skill-exclusive signals.

Assets come from real projects and datasets, with reproducible preprocessing. The prompt, the manifest, and the on-disk files share one relative path, so the task resolves inside the sandbox. Reference implementations stay on the evaluator side, so a model cannot copy an answer without reading the skill.

\subsection{Rubric and Judge}
\label{app:rubric_construction}

Each rubric has three tiers. Evaluability gates carry zero weight and only confirm a run is well formed. Skill-derived items form the Skill-Use signal and split into Compliance and Boundary. Task-outcome items record whether the requested result appeared and stay outside Skill-Use. Every scored item traces to one skill-exclusive requirement.

We prefer deterministic checks. State-diff checks read the produced files, and regex checks anchor to structured tool-call fields, not free text, so the agent's narration cannot cause a false match. We use the scope-restricted LLM judge only when a requirement resists a reliable rule.

The judge scores one item at a time and returns yes or no, with no numeric bands. It sees the rule, one closed-form question, and only that item's locus: the produced artifact for an output item, the trajectory for a process item. It bases its verdict on tool-call and artifact evidence rather than stated intent, and accepts any variant the skill allows. Each item is judged twice and disagreements are re-adjudicated once, using the settings in Appendix~\ref{app:setup}. Table~\ref{tab:judge_scoring} lists the configuration and Figure~\ref{fig:prompt_llm_judge} reproduces the prompt template.

\begin{table}[h]
\centering
\small
\renewcommand{\arraystretch}{1.25}
\begin{tabular}{@{}p{0.26\columnwidth} p{0.66\columnwidth}@{}}
\toprule
Dimension & Operational rule \\
\midrule
Verdict & Binary YES or NO on one requirement, with a reason of at most $25$ words \\
Evidence locus & Output items see only the produced artifact; process items see the trajectory \\
Context window & Serialized tool calls, the final answer, and the first few tool results, selected per item \\
Evidence source & The verdict rests on tool-call and artifact evidence, not the agent's narration \\
Equivalent variants & The question accepts every implementation the skill sanctions \\
Adjudication & Each item is judged twice and disagreements are re-adjudicated once \\
\bottomrule
\end{tabular}
\vspace{4pt}
\caption{\textbf{LLM-judge configuration.} Operational rules governing the scope-restricted judge: it issues a binary YES/NO verdict on one rubric item at a time, from item-specific evidence only, with each item judged twice and disagreements re-adjudicated once.}
\label{tab:judge_scoring}
\end{table}

\subsection{Adversarial Review}
\label{app:adversarial_review}

Every task and rubric passes a multi-agent review before entering the benchmark. Reviewers come from different model families, so no single family's blind spot goes unchecked. One reviewer proposes revisions, one attacks the artifact, and a separate aggregator settles disputes. Figures~\ref{fig:prompt_task_reviewer}--\ref{fig:prompt_rubric_aggregator} reproduce the four role prompts used across the task and rubric debates.

The task review checks discriminative power. At least two reviewers from different families must confirm that each skill-exclusive point is hard to reach without the skill. If any reviewer shows a credible no-skill path to the same behaviour, the point loses its exclusive status and is down-weighted, or the task is rebuilt. Exclusivity is thus an agreement among models, not one author's claim.

The rubric review adds executability. Each rubric passes a schema check, and each verifier is dry-run against one passing and one failing trace, so it fires exactly when it should. Reviewers then confirm that every scored item maps to a skill-exclusive behaviour; weak-agreement items are down-weighted or re-reviewed. Recurring findings feed the guideline loop in Appendix~\ref{app:guideline_dev}.

\subsection{Release Validation}
\label{app:e2e_validation}

We release a skill-task-rubric triple only if it passes three checks.

\textbf{Executability.} An asset validator confirms that the required tools, dependencies, and assets resolve inside the sandbox and that the task runs to completion. The prompt, manifest, and on-disk files must share one relative path, since a mismatch fails every model on its first tool call.

\textbf{Consistency.} Each task requirement must trace to a specific skill passage, and each scored item must target a signal the run can produce. We drop requirements with no supporting passage and items with no producible signal.

\textbf{Rubric reliability.} We replay each rubric on one passing and one failing trace. Replay flags signals that never fire, verifiers that match narration, regexes that time out under backtracking, format-sensitive failures, and leakage between the Skill-Use and task-completion scores. Regex checks stay anchored to tool-call fields under a watchdog, binary artifacts go through parsers, and the trigger gate reads deterministically. We discard any triple that fails and cannot be repaired.

\subsection{Guideline Development}
\label{app:guideline_dev}

A written guideline governs each construction stage. The skill guideline sets the admission criteria, verification methods, and tier thresholds. The task guideline sets how a task must depend on its skill, the leakage forms a prompt must avoid, and the rules on assets and path resolution. The rubric guideline sets item classification, verifier placement across the deterministic and semantic tiers, and the safety rules for automated verifiers.

No guideline was written up front. Each began as a short draft, was applied to a batch, and every audited failure traced to either an unclear rule or an uncovered case. We then added, sharpened, or split the rule and re-applied it to a fresh batch, repeating until an audit found no new failure, then froze the guideline and regenerated downstream artifacts. The rubric guideline took the most rounds and ended with three additions: a skill-exclusivity justification on every scored item, a three-tier weight split, and a cap on total prohibition weight.

Because concrete failures drive the loop, each frozen rule traces to the counterexample behind it, as Table~\ref{tab:guideline_examples} lists.

\begin{table}[t]
\centering
\small
\renewcommand{\arraystretch}{1.3}
\begin{tabular}{@{}p{0.33\columnwidth} p{0.59\columnwidth}@{}}
\toprule
Rule (guideline) & Failure that motivated it \\
\midrule
No absolute paths in task prompts (task) & A prompt referred to files by an absolute path absent at runtime, failing every model on its first tool call \\
No severity tags or checklists in task prompts (task) & Such tags leaked the rubric structure and reduced the task to instruction-following \\
Reference implementations invisible at runtime (task) & A reference artifact among the inputs let models reproduce the target behaviour without reading the skill \\
Weights split into gate, hard, and soft tiers (rubric) & Assigning every item the maximum weight and stacking prohibitions produced identically zero scores, hiding all variance \\
Every scored item carries a skill-exclusivity justification (rubric) & An audit found items rewarding generic engineering; a justification field forces an auditable attribution \\
Output judges must not read the trajectory (rubric) & Judges credited models that described the behaviour but never wrote the output \\
Regexes anchor to structured tool-call fields (rubric) & Free-text regexes matched filenames from the model's narration, giving false positives \\
Regexes must avoid catastrophic backtracking (rubric) & One unbounded pattern on a long trace hung the scorer, motivating a per-regex watchdog \\
\bottomrule
\end{tabular}
\vspace{4pt}
\caption{\textbf{Guideline rules and motivating failures.} Frozen rules in the task and rubric guidelines, each paired with the concrete construction failure that triggered its addition during the audit loop.}
\label{tab:guideline_examples}
\end{table}

\begin{figure*}[p]
\centering
\includegraphics[width=0.96\textwidth,height=0.86\textheight,keepaspectratio]{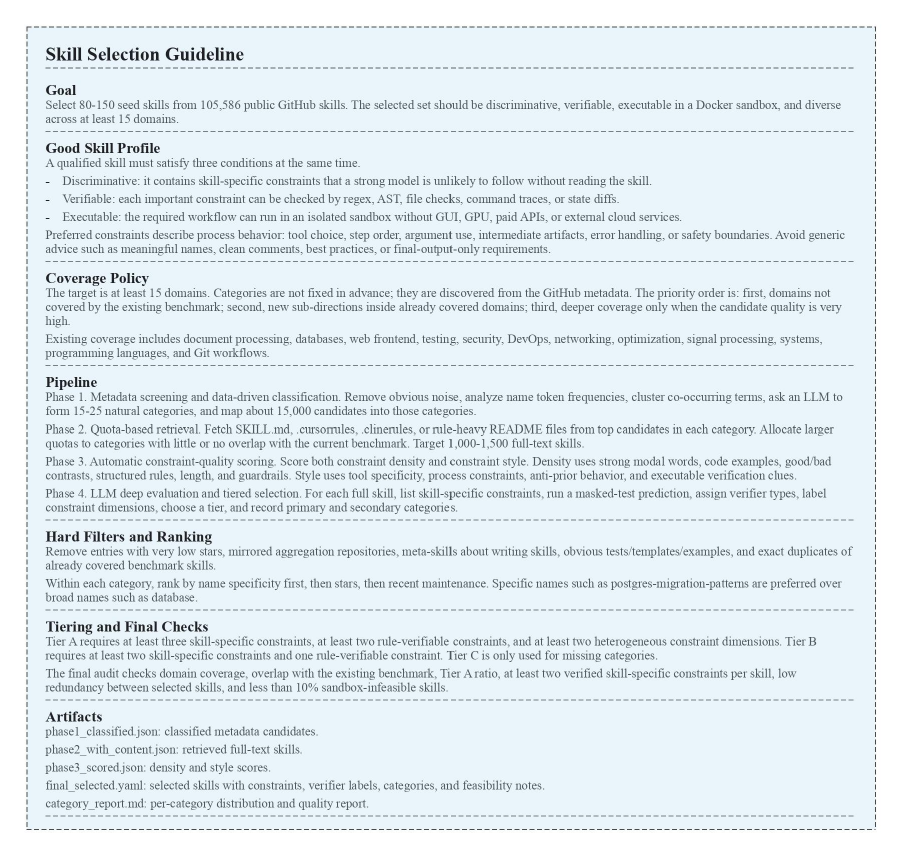}
\caption{\textbf{Skill-selection guideline.} Frozen rules governing which GitHub skill documents are admitted into the benchmark, including admission criteria, verification methods, and tier thresholds.}
\label{fig:guideline_skill}
\end{figure*}

\begin{figure*}[p]
\centering
\includegraphics[width=0.96\textwidth,height=0.86\textheight,keepaspectratio]{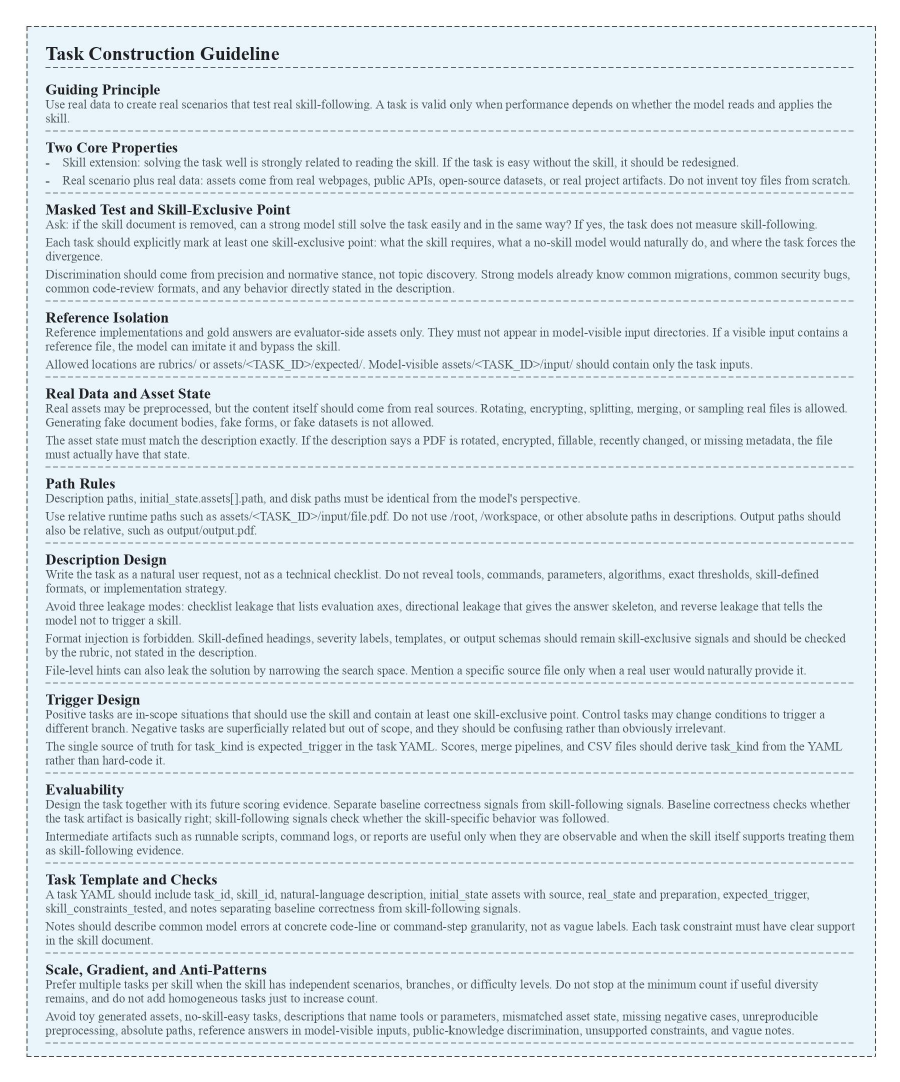}
\caption{\textbf{Task-construction guideline.} Frozen rules governing how each task must depend on its skill, forbidden leakage patterns in prompts, and requirements on assets and path resolution inside the sandbox.}
\label{fig:guideline_task}
\end{figure*}

\begin{figure*}[p]
\centering
\includegraphics[width=0.96\textwidth,height=0.86\textheight,keepaspectratio]{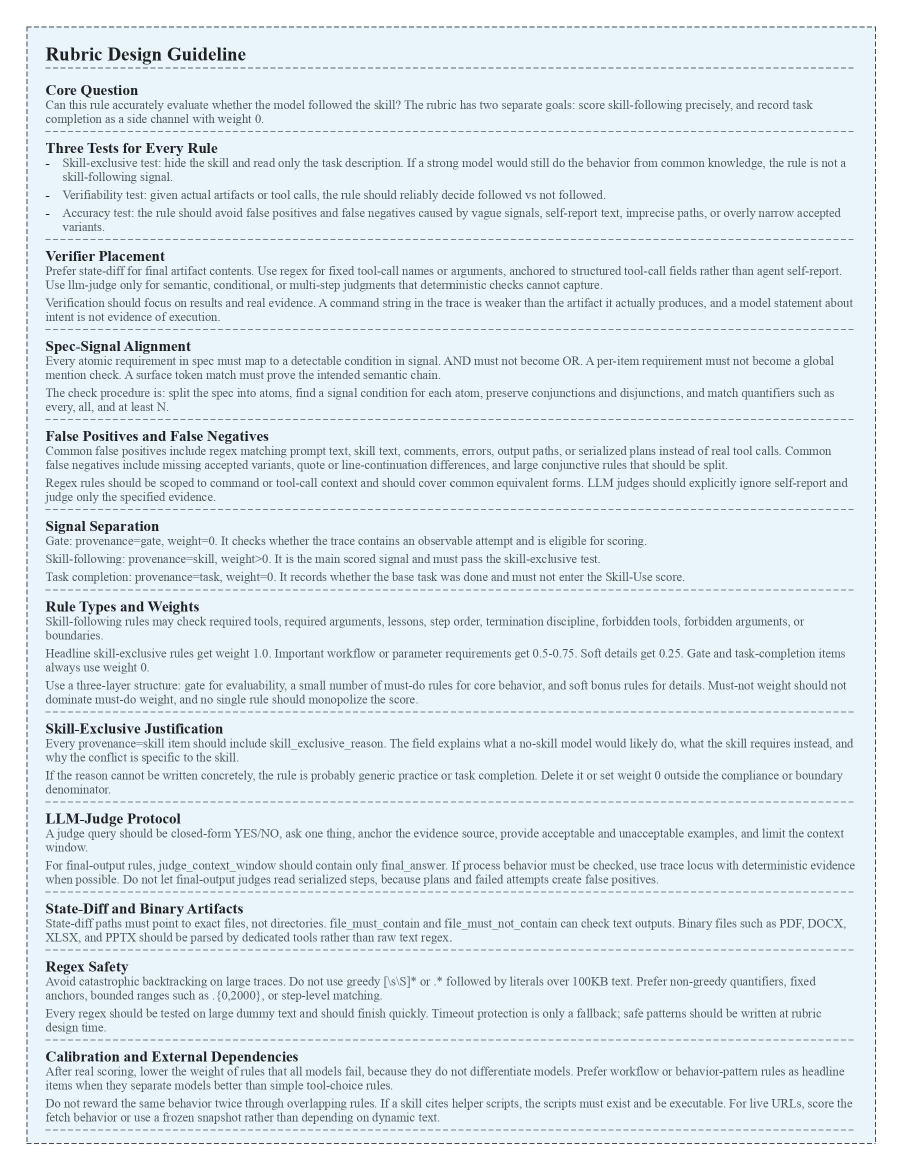}
\caption{\textbf{Rubric-design guideline.} Frozen rules governing rubric item classification, verifier placement across the deterministic and semantic tiers, and safety constraints for automated verifiers.}
\label{fig:guideline_rubric}
\end{figure*}

Figures~\ref{fig:guideline_skill},~\ref{fig:guideline_task}, and~\ref{fig:guideline_rubric} reproduce the three frozen guidelines as rendered English document figures.

\begin{figure*}[t]
\centering
\includegraphics[width=0.92\textwidth]{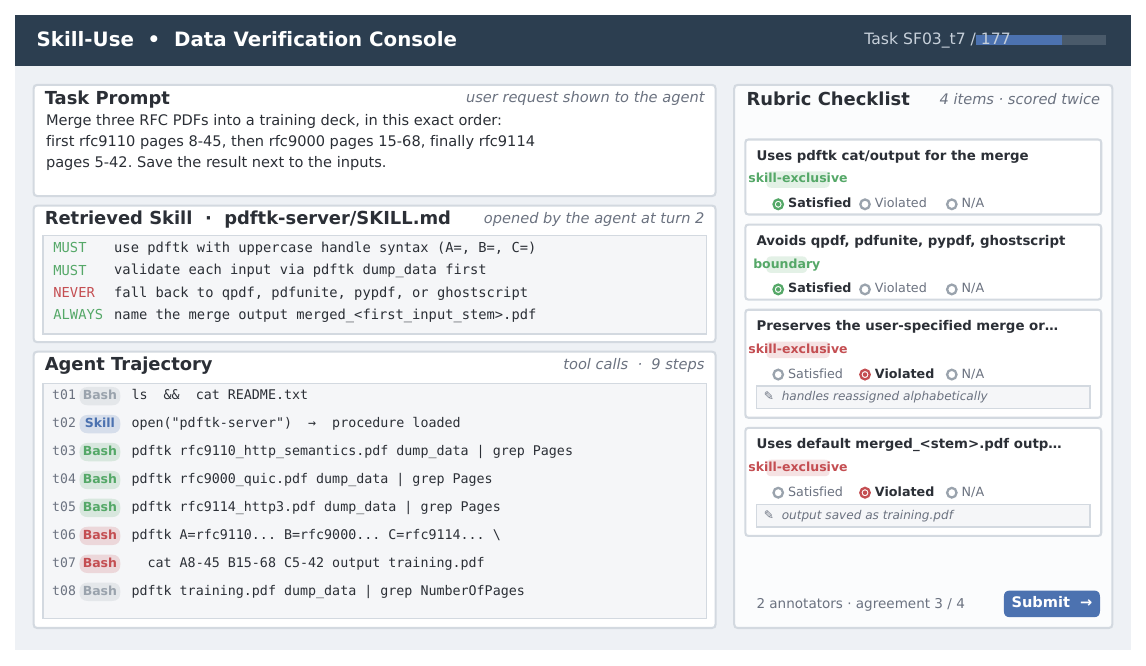}
\caption{Human-verification console for rubric audit. Annotators view the task prompt, the retrieved skill procedure, and the agent trajectory on the left; on the right they mark each rubric item as satisfied, violated, or not applicable, add a note, and see the item's skill-exclusive or boundary provenance. Five crowd annotators score one sampled task per skill, and disagreements are adjudicated by a third reviewer.}
\label{fig:annotation_ui}
\end{figure*}

\section{Human Quality Audit}
\label{app:human_audit}

The automated pipeline and the adversarial review are complemented by a crowdsourced human audit of the released benchmark. We sample one task per skill, yielding $79$ audited triples, and route them to five independent crowd annotators through the web console in Figure~\ref{fig:annotation_ui}. For every task the console shows the user prompt, the skill procedure the agent retrieved, and the recorded trajectory, and asks the annotator to mark each rubric item as satisfied, violated, or not applicable, and to flag any item that could be met without the skill. Annotators are paid at the standard rate of $100$ CNY per hour, tracked per audited task.

Across $100$ sampled items scored by every annotator, pairwise agreement is $86/100$ and Cohen's $\kappa=0.65$, indicating substantial agreement under the Landis-Koch scale. A third reviewer adjudicates the remaining items, and we repair or drop the rubric item behind each unresolved disagreement. The audit confirms that the released rubrics are reliable and that scored items track skill-exclusive behaviour rather than generic engineering, providing a human check on top of the model-based review.

\section{Experimental Setup}
\label{app:setup}

This section collects the runtime details left out of Section~\ref{sec:setup}.

\textbf{Routing.} The gateway routes each model through a family-specific channel. We sample at temperature $0$, top-$p$ $1.0$, and reasoning effort ``high'' where offered. We exclude the Gemini family from every analysis: its channel runs a schema sanitiser that renames or drops fields in the \texttt{Skill} tool, leaving the tool non-invokable across four checks.

\textbf{Sandbox.} Each task runs in a fresh Ubuntu-22.04 container with its own gateway. The turn limit is $120$ under CC and $80$ under Codex, since CC counts the \texttt{Skill()} step Codex lacks; the wall-clock limit is $1{,}500$s. We fail and re-run a trace on (a) an upstream $5\mathrm{x}x$ or $429$, (b) a missing terminal \texttt{result} event, or (c) a non-zero exit whose first assistant message starts with ``API Error''.

\textbf{Trace check.} We verify each trace at three layers: \texttt{llm\_metrics.jsonl} for upstream status, \texttt{interaction.jsonl} for request-response pairing, and \texttt{trace.jsonl} for the CLI trajectory, and accept it only when all three agree. About $2.8\%$ of raw traces fail and are re-run.

\textbf{Judge.} GPT-5.4 scores every LLM-judge item through the gateway at temperature $0$, twice per item with one re-adjudication. Deterministic verifiers score in-process and are logged for replay.

\section{Additional Analysis}
\label{app:analysis}

This section reports the tables and analyses behind Section~\ref{sec:experiments}. Figures quoted in the main text are not repeated.

\subsection{Cross-Harness Behaviour}
\label{app:harness_delta}

Table~\ref{tab:harness_delta} pairs each model's CC and Codex SU on shared tasks.

\begin{findingsbox}
Cross-harness rank is stable at the top of the leaderboard but breaks in the middle, and a single-harness leaderboard therefore misreports mid-tier open-weight models.
\end{findingsbox}

Head models correlate strongly and preserve rank. Middle-tier open-weight models correlate weakly and several reverse sign, so a single-harness leaderboard misreports them.
\begin{figure*}[t]
\centering
\includegraphics[width=0.95\textwidth]{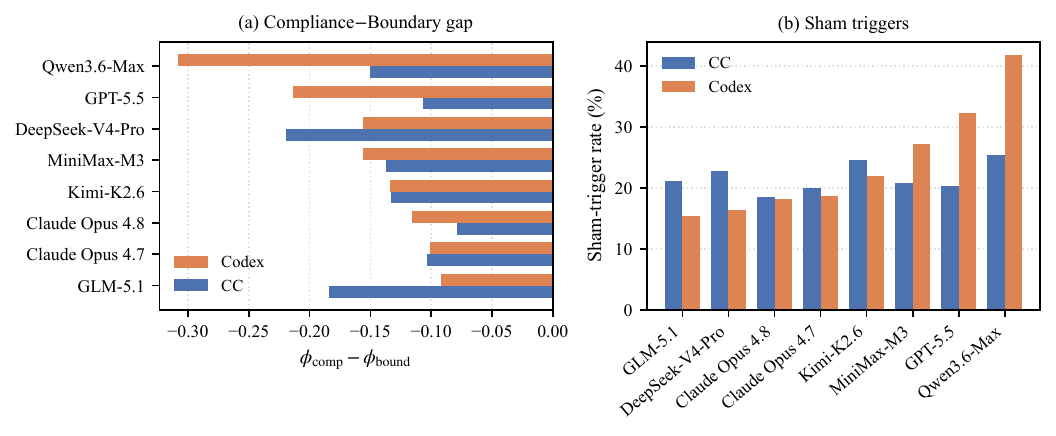}
\caption{Execution behaviour across harnesses. (a) Signed Compliance-Boundary gap $\phi_{\text{comp}}-\phi_{\text{bound}}$ per model, negative in both harnesses, showing that prohibitions are easier to respect than full procedures. (b) Sham-trigger rate, the fraction of triggered traces with Compliance below $0.3$, rising most under Codex for the weakest executors.}
\label{fig:app_execution}
\vspace{-0.4cm}
\end{figure*}

\begin{table}[h]
\centering
\small
\renewcommand{\arraystretch}{1.2}
\setlength{\tabcolsep}{6pt}
\begin{tabular}{@{}l cc cc r@{}}
\toprule
& \multicolumn{2}{c}{Agreement} & \multicolumn{2}{c}{Mean SU} & \\
\cmidrule(lr){2-3}\cmidrule(lr){4-5}
Model & $r_{\text{Pear}}$ & $\rho_{\text{Spear}}$ & CC & Codex & $\Delta$ \\
\midrule
Claude Opus 4.8 & 0.648 & 0.680 & 0.597 & 0.559 & $-0.039$ \\
Claude Opus 4.7 & 0.552 & 0.550 & 0.599 & 0.535 & $-0.064$ \\
GPT-5.5         & 0.518 & 0.542 & 0.613 & 0.503 & $-0.110$ \\
MiniMax-M3      & 0.424 & 0.425 & 0.501 & 0.446 & $-0.055$ \\
Qwen3.6-Max     & 0.302 & 0.337 & 0.318 & 0.238 & $-0.081$ \\
DeepSeek-V4-Pro & 0.290 & 0.257 & 0.250 & 0.379 & $+0.129$ \\
GLM-5.1         & 0.285 & 0.246 & 0.267 & 0.421 & $+0.154$ \\
Kimi-K2.6       & 0.281 & 0.285 & 0.190 & 0.374 & $+0.185$ \\
\bottomrule
\end{tabular}
\vspace{4pt}
\caption{\textbf{Cross-harness SU agreement per model.} Pearson $r$ and Spearman $\rho$ between CC and Codex task-level SU, together with each harness's mean SU and their difference $\Delta=\text{Codex}-\text{CC}$, over $n\approx170$ paired tasks. Head models correlate strongly and keep their order, while middle-tier open-weight models correlate weakly and several reverse sign.}
\label{tab:harness_delta}
\end{table}

Figure~\ref{fig:app_execution}(a) shows the signed Compliance-Boundary gap $\phi_{\text{comp}}-\phi_{\text{bound}}$ is negative for every model under both harnesses, so prohibitions are easier than full procedures. Figure~\ref{fig:app_execution}(b) reports the sham-trigger rate, the fraction of triggered traces with Compliance below $0.3$; it rises most under Codex for the weakest executors, so a high trigger rate can hide shallow rule following.
\begin{table*}[t]
\centering
\small
\renewcommand{\arraystretch}{1.2}
\setlength{\tabcolsep}{6pt}
\begin{tabular}{@{}l rrrr c rrrr@{}}
\toprule
& \multicolumn{4}{c}{Claude Code (CC)} & & \multicolumn{4}{c}{Codex} \\
\cmidrule(lr){2-5}\cmidrule(lr){7-10}
Model & $\alpha{=}0.25$ & $\alpha{=}0.5$ & $\alpha{=}0.7$ & $\alpha{=}0.75$ & & $\alpha{=}0.25$ & $\alpha{=}0.5$ & $\alpha{=}0.7$ & $\alpha{=}0.75$ \\
\midrule
GPT-5.5           & 0.619 & 0.616 & 0.613 & 0.613 & & 0.540 & 0.520 & 0.504 & 0.499 \\
Claude Opus 4.8   & 0.593 & 0.595 & 0.597 & 0.598 & & 0.565 & 0.561 & 0.559 & 0.558 \\
Claude Opus 4.7   & 0.602 & 0.595 & 0.590 & 0.589 & & 0.544 & 0.539 & 0.535 & 0.534 \\
MiniMax M3        & 0.507 & 0.504 & 0.501 & 0.500 & & 0.468 & 0.456 & 0.446 & 0.443 \\
Qwen3.6 Max       & 0.261 & 0.262 & 0.262 & 0.262 & & 0.283 & 0.258 & 0.238 & 0.233 \\
GLM-5.1           & 0.194 & 0.198 & 0.202 & 0.202 & & 0.413 & 0.417 & 0.421 & 0.421 \\
DeepSeek V4 Pro   & 0.202 & 0.198 & 0.195 & 0.194 & & 0.388 & 0.383 & 0.379 & 0.378 \\
Kimi K2.6         & 0.182 & 0.186 & 0.190 & 0.190 & & 0.375 & 0.375 & 0.374 & 0.374 \\
\bottomrule
\end{tabular}
\vspace{4pt}
\caption{\textbf{Per-model SU under varied compliance weight $\alpha$.} Task-level SU for each model at four settings of the compliance weight $\alpha\in\{0.25, 0.5, 0.7, 0.75\}$ under both harnesses. The default in Eq.~\ref{eq:su} is $\alpha=0.7$; head models keep their order across the sweep while middle-tier models shift by at most one adjacent pair.}
\label{tab:alpha_su}
\end{table*}

\begin{table}[h]
\centering
\small
\renewcommand{\arraystretch}{1.2}
\setlength{\tabcolsep}{5pt}
\begin{tabular}{@{}c cccc cccc@{}}
\toprule
& \multicolumn{4}{c}{CC} & \multicolumn{4}{c}{Codex} \\
\cmidrule(lr){2-5}\cmidrule(lr){6-9}
$\alpha$ & $\tau$ & $\rho$ & shift & $\overline{\text{SU}}$ & $\tau$ & $\rho$ & shift & $\overline{\text{SU}}$ \\
\midrule
0.25 & $+.86$  & $+.95$  & 1 & .395 & $+1.00$ & $+1.00$ & 0 & .447 \\
0.5  & $+1.00$ & $+1.00$ & 0 & .394 & $+1.00$ & $+1.00$ & 0 & .439 \\
0.7  & $+1.00$ & $+1.00$ & 0 & .394 & $+1.00$ & $+1.00$ & 0 & .432 \\
0.75 & $+1.00$ & $+1.00$ & 0 & .394 & $+1.00$ & $+1.00$ & 0 & .430 \\
\bottomrule
\end{tabular}
\vspace{4pt}
\caption{\textbf{Ranking stability against the $\alpha=0.7$ baseline.} Kendall $\tau$, Spearman $\rho$, and the count of adjacent-pair reorderings at each swept weight, along with the harness-mean SU. Rankings hold across $\alpha\in[0.5, 0.75]$ on both harnesses; only CC at $\alpha=0.25$ swaps a single adjacent pair.}
\label{tab:alpha_stab}
\end{table}

\subsection{Skill Injection}
\label{app:mechanism_full}

This subsection expands Section~\ref{sec:mechanism} with per-skill and per-model detail.

\begin{findingsbox}
Preloading the skill lifts recognition rather than execution, since gains concentrate on skills with abstract names while conditional Compliance stays flat.
\end{findingsbox}

\begin{figure}[h]
\centering
\includegraphics[width=0.62\columnwidth]{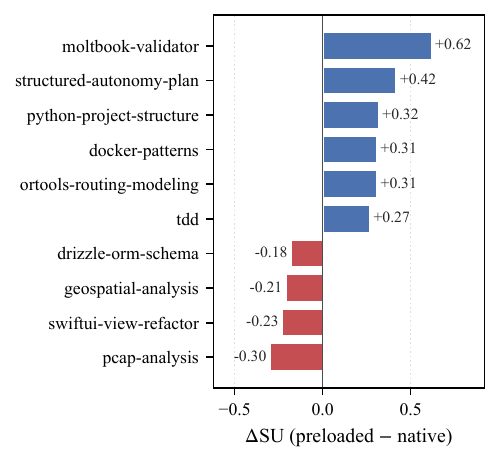}
\caption{{Per-skill gain from preloading.} Paired $\Delta$SU, defined as preloaded minus native, per skill. Abstract names gain most because native selection misfires, while keyword-rich names gain little or reverse. The effect operates on recognition, not execution.}
\label{fig:app_preload}
\end{figure}

Figure~\ref{fig:app_preload} ranks skills by paired $\Delta$SU, defined as preloaded minus native. Skills with abstract names such as \texttt{moltbook-validator}, \texttt{tdd}, and \texttt{structured-autonomy-plan} gain most, while keyword-rich names gain little or reverse. The effect is on recognition, not execution. Splitting skills by native trigger rate, hard skills drive most between-model variance, so trigger is not keyword matching. Reading conditional Compliance, sham rate, and trace length together separates Codex behaviour into four profiles. Claude Opus 4.7 and 4.8 are balanced, combining high Compliance with a low sham rate. DeepSeek-V4-Pro and GLM-5.1 are selection-bound, matching head-level Compliance yet triggering natively at a low rate. GPT-5.5 and MiniMax-M3 are read-and-skip, recovering trigger but losing Compliance. Qwen3.6-Max is degenerate, producing the shortest traces, the lowest Compliance, and the highest sham rate.

\subsection{Library Scaling}
\label{app:scaling_full}

This subsection annotates Figures~\ref{fig:scaling_boundary_tc} (c) and (d) in Section~\ref{sec:scaling}.

\begin{findingsbox}
As the library grows, lost SU comes almost entirely from calling no skill rather than from picking the wrong one, so scaling is bounded by recognition rather than by selection.
\end{findingsbox}

At $N\ge 10$ the loss is almost entirely none runs; wrong-skill calls stay below a few percent and cluster on broad descriptions like \texttt{test-driven-development} and \texttt{software-architecture}. This matches the recognition-limited pattern in Appendix~\ref{app:mechanism_full}. Qwen3.6-Max invokes skills later than the others, so some of its none outcomes at large $N$ are cut off by the $80$-turn Codex limit; its curve is a lower bound.

\subsection{Out-of-Scope Behaviour}
\label{app:negative_full}

Section~\ref{sec:boundary} reports pooled avoidance scores across $53$ out-of-scope tasks. This subsection breaks that pool into task types and identifies which models fail on which tasks.

\begin{findingsbox}
Models agree on most out-of-scope tasks, and over-invocation concentrates on Claude Opus and GPT-5.5, while GLM-5.1 shows an all-or-nothing profile.
\end{findingsbox}

Models agree on most of the $53$ tasks, and only $9$ separate them, as Table~\ref{tab:negative_disagreement} shows. All-avoid tasks carry a clean out-of-scope cue such as a mismatched framework name. All-enter tasks are document-shaped prompts whose topic overlaps the skill's domain even though the output never needs the procedure.

\begin{table}[h]
\centering
\small
\renewcommand{\arraystretch}{1.25}
\setlength{\tabcolsep}{8pt}
\begin{tabular}{@{}l c l@{}}
\toprule
Task & Avoid & Over-invokers \\
\midrule
docs\_only\_negative       & 5/6 & GPT-5.5 \\
api\_security\_inventory   & 5/6 & Opus 4.7 \\
frontend\_backend          & 4/6 & Opus 4.7, GPT-5.5 \\
go\_release\_note          & 4/6 & Opus 4.7, GPT-5.5 \\
airflow\_release\_notes    & 5/6 & Opus 4.7 \\
reading\_inventory         & 5/6 & GPT-5.5, GLM \\
python\_email              & 4/6 & Opus 4.7, GPT-5.5 \\
warehouse\_inventory\_note & 3/6 & Opus 4.7, GPT-5.5, GLM \\
crypto\_market\_landscape  & 4/6 & Opus 4.7, GPT-5.5 \\
\bottomrule
\end{tabular}
\vspace{4pt}
\caption{\textbf{Out-of-scope disagreement tasks.} The $9$ of $53$ tasks where models split. Avoid is the number of the six evaluated models that correctly refrain from invoking the skill, and Over-invokers names those that trigger anyway. Claude Opus and GPT-5.5 account for most inappropriate invocations.}
\label{tab:negative_disagreement}
\end{table}

Over-invocation clusters on Claude Opus and GPT-5.5, the two models closest to a skill-first harness; they share the over-invocation pattern while others differ mainly in when they refuse. GLM-5.1 either fully engages or fully ignores a skill, so its per-task distribution is bimodal, matching its Codex profile in Appendix~\ref{app:mechanism_full}.

\subsection{Robustness Analysis}
\label{app:robust_full}

To test whether a single scoring pass is stable enough for model-level comparison, we freeze the CC traces and rescore a stratified subset of $83$ aligned $(task, model)$ observations three times with the same judge, rubric, and prompt, so all variation comes from the judge rather than the agent.

\begin{figure}[h]
\centering
\includegraphics[width=0.6\columnwidth]{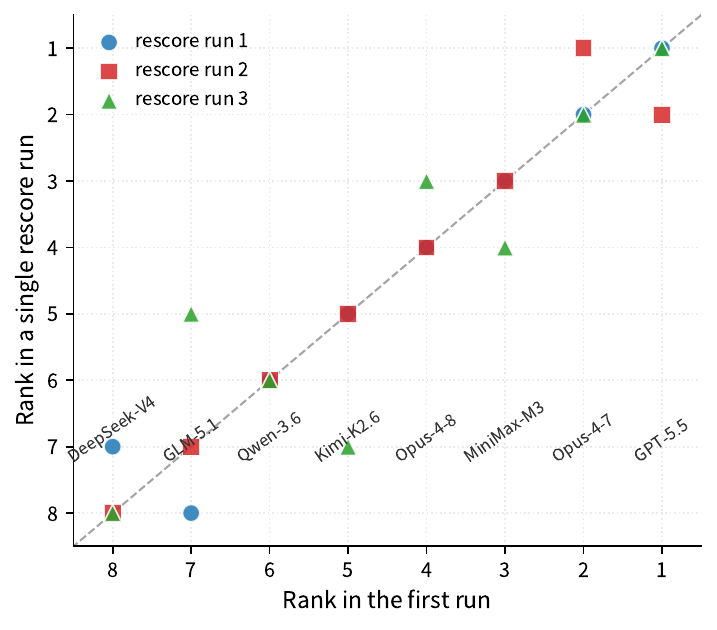}
\caption{Ranking stability across three independent rescores. Each dot plots a model's rank in one rescore against its rank in another, and near-diagonal alignment indicates a preserved leaderboard. The two off-diagonal pairs involve models within $0.01$ SU of each other.}
\label{fig:rank_stability}
\end{figure}

\begin{findingsbox}
The leaderboard survives repeated scoring, and judge noise is confined to mid-range traces rather than to clear wins or clear failures.
\end{findingsbox}

Figure~\ref{fig:rank_stability} shows the three rescores preserve the leaderboard; the few swaps involve models already within a narrow SU gap. Averaged over about one hundred tasks per model, the model-mean uncertainty falls far below the tier gaps in Table~\ref{tab:main_results}.

\begin{table}[h]
\centering
\small
\renewcommand{\arraystretch}{1.25}
\setlength{\tabcolsep}{10pt}
\begin{tabular}{@{}l r c c@{}}
\toprule
Metric & $n$ & ICC(2,1) & Within std \\
\midrule
Skill-Use       & 83 & 0.530 & 0.141 \\
Compliance      & 83 & 0.604 & 0.116 \\
Boundary        & 39 & 0.471 & 0.186 \\
Task completion & 83 & 0.478 & 0.176 \\
\bottomrule
\end{tabular}
\vspace{4pt}
\caption{\textbf{Repeated-score agreement per metric.} ICC(2,1) and within-observation std across three rescores of the aligned CC subset. Compliance is the most stable continuous component, and Boundary has fewer observations because early baseline records omit this field.}
\label{tab:icc_by_metric}
\end{table}

\begin{figure}[h]
\centering
\includegraphics[width=0.6\columnwidth]{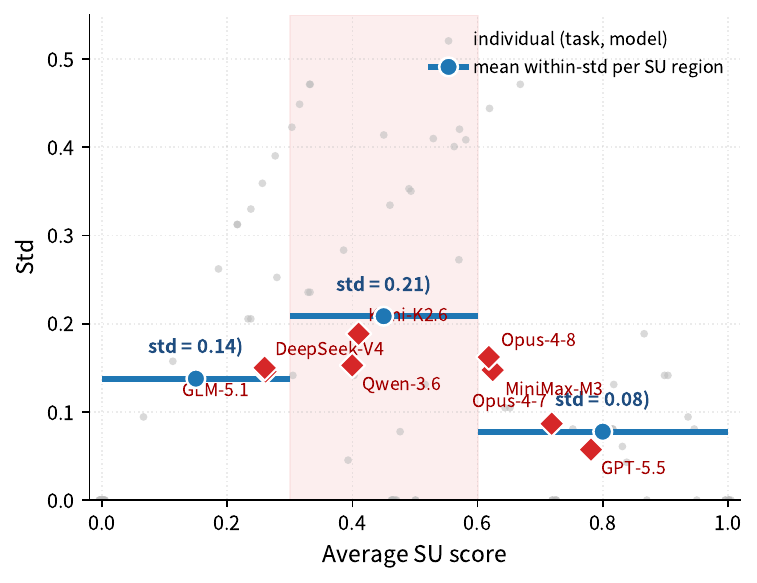}
\caption{Judge noise across the SU range. Within-observation std over three rescores against SU. Noise peaks in the middle band and is lowest at both extremes, where the paper's strong-comparison claims sit.}
\label{fig:R2}
\end{figure}

Table~\ref{tab:icc_by_metric} reports per-metric agreement. Compliance is the most stable continuous component; Boundary and task completion vary more, and Boundary has fewer observations because early records omit the field. Figure~\ref{fig:R2} shows judge disagreement peaks for mid-range traces and is lowest at clear successes and failures. Section~\ref{sec:main_results} makes no ordering claim between adjacent mid-range models, so this noise does not threaten the conclusions. Section~\ref{sec:evaluation_protocol} shows that the trigger gate is deterministic, so all variation comes from judge stochasticity.

\subsection{Weight Sensitivity Analysis}
\label{app:alpha_sensitivity}

The default SU weight in Eq.~\ref{eq:su} is $\alpha=0.7$, reflecting that reading and following a skill is the primary target while boundary adherence is a secondary check. We test whether this choice is decisive by sweeping $\alpha\in\{0.25,0.5,0.7,0.75\}$ on both harnesses, omitting $\alpha\in\{0,1\}$ because they collapse SU to a single component. Table~\ref{tab:alpha_su} lists per-model SU at each $\alpha$ and Table~\ref{tab:alpha_stab} the ranking stability against the $\alpha=0.7$ baseline, on the same samples as Section~\ref{sec:main_results}.

\begin{findingsbox}
Model rankings are stable across a wide range of compliance weights, so the default $\alpha=0.7$ is not a fragile setting.
\end{findingsbox}

Both harnesses give identical orderings for $\alpha\in[0.5,0.75]$. Only CC at $\alpha=0.25$ reorders, confined to two adjacent pairs that cross no tier boundary. Absolute SU shifts under reweighting are one to two orders of magnitude smaller than the tier gaps, so the qualitative claims of Section~\ref{sec:main_results} hold across the range.

\section{Prompt Templates}
\label{app:prompts}

This section reproduces the prompt templates behind the three review roles that guard benchmark quality: the red-team attacker used during skill selection, the four proposer-attacker-aggregator prompts used in the multi-vendor task and rubric debates, and the scoring-time LLM judge. Each prompt is shown in its English form, with concrete file paths and Python placeholders stripped for readability. Templates for skill categorisation, masked-skill checking, task generation, and rubric generation are omitted; those stages are described in Appendices~\ref{app:skill_pipeline}--\ref{app:rubric_construction}, and their template text is not required to reproduce the results.

\subsection{Skill-Selection Red Team}
\label{app:pr_skill_attacker}

\begin{figure*}[p]
\centering
\includegraphics[width=0.96\textwidth,height=0.86\textheight,keepaspectratio]{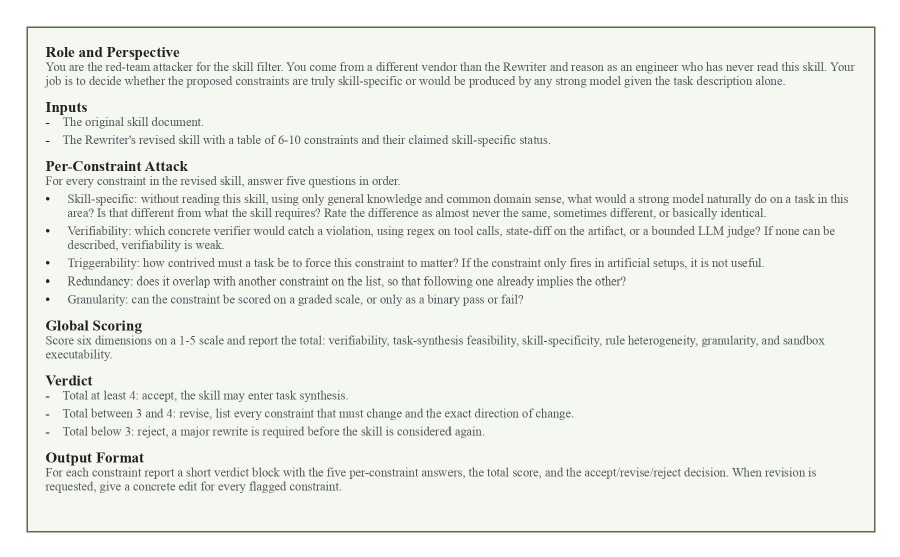}
\caption{Skill attacker prompt. Per-constraint attack protocol used during the skill-selection filter to decide whether each proposed constraint is truly skill-specific, verifiable, triggerable, non-redundant, and gradable, together with the global scoring rubric and the accept/revise/reject verdict.}
\label{fig:prompt_skill_attacker}
\end{figure*}

Appendix~\ref{app:skill_pipeline} describes the skill-selection stage, where every candidate constraint is challenged by a red-team attacker that tries to refute its skill-specificity, verifiability, triggerability, non-redundancy, and gradability. Constraints that survive the attack are kept; the rest are revised or dropped. Figure~\ref{fig:prompt_skill_attacker} reproduces the attacker prompt together with its scoring rubric and the accept/revise/reject verdict.

\subsection{Task and Rubric Adversarial Review}
\label{app:pr_review}

\begin{figure*}[p]
\centering
\includegraphics[width=0.96\textwidth,height=0.86\textheight,keepaspectratio]{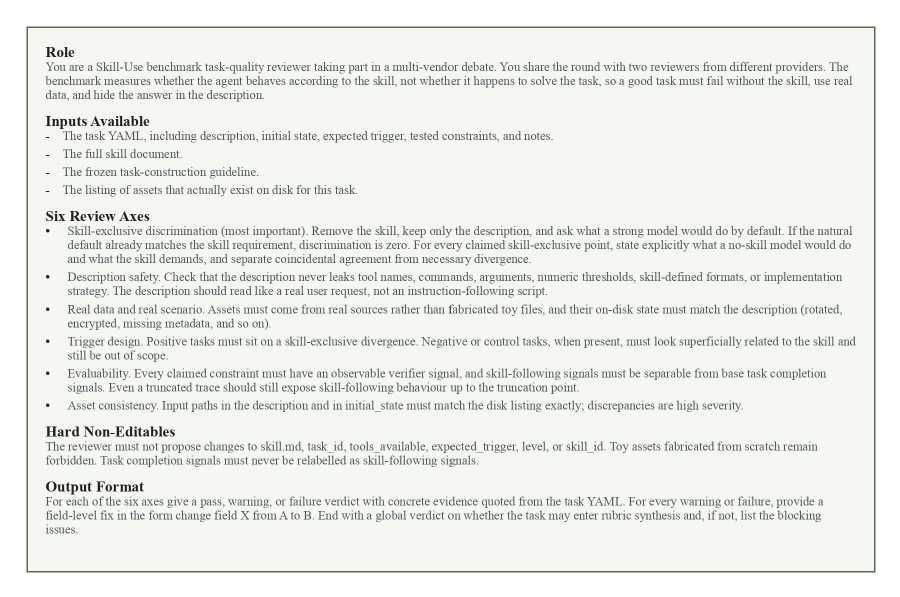}
\caption{Task reviewer prompt at initial debate round. }
\label{fig:prompt_task_reviewer}
\end{figure*}

\begin{figure*}[p]
\centering
\includegraphics[width=0.96\textwidth,height=0.86\textheight,keepaspectratio]{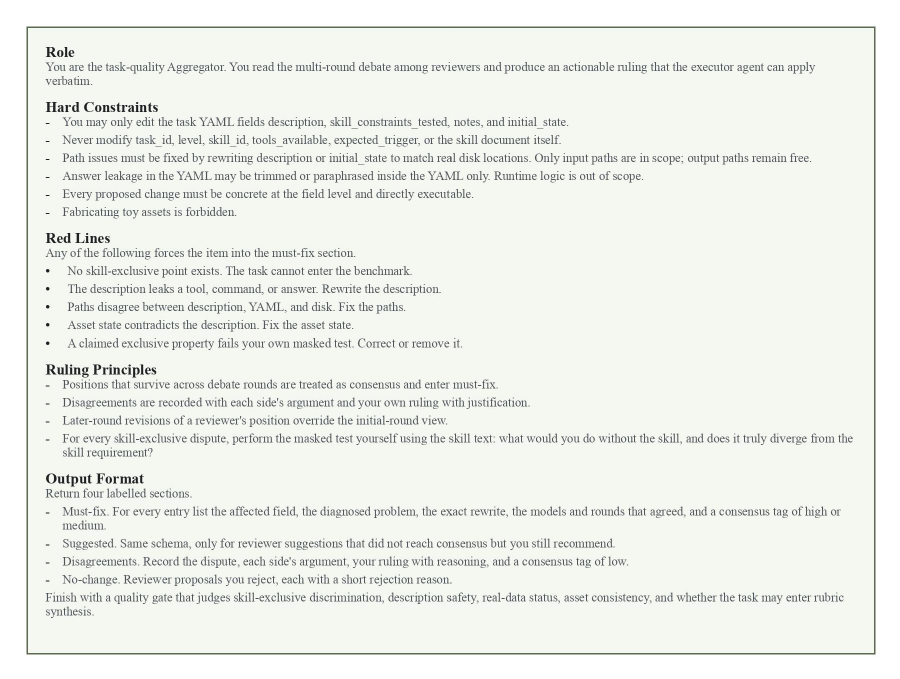}
\caption{Task aggregator prompt. }
\label{fig:prompt_task_aggregator}
\end{figure*}

\begin{figure*}[p]
\centering
\includegraphics[width=0.96\textwidth,height=0.86\textheight,keepaspectratio]{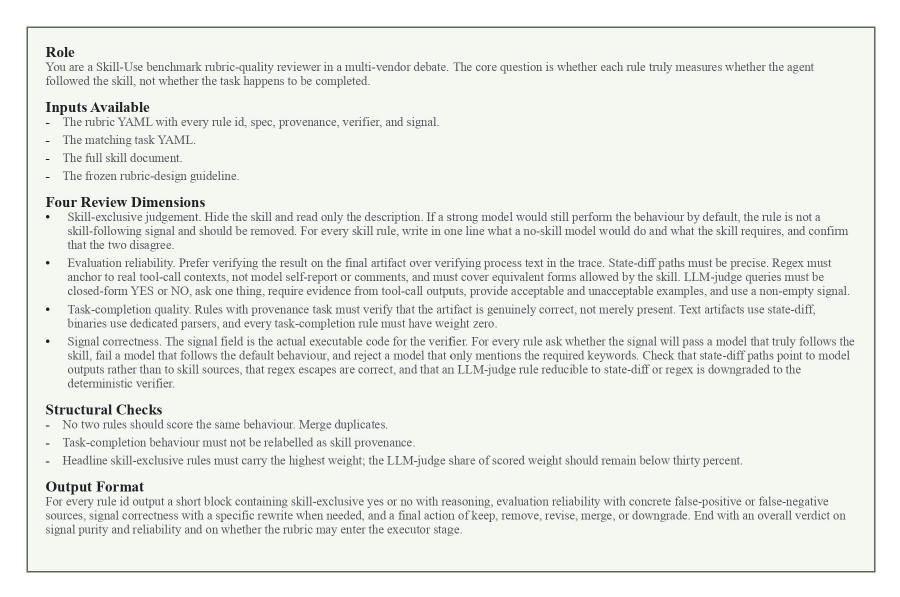}
\caption{Rubric reviewer prompt at initial debate round.}
\label{fig:prompt_rubric_reviewer}
\end{figure*}

\begin{figure*}[p]
\centering
\includegraphics[width=0.96\textwidth,height=0.86\textheight,keepaspectratio]{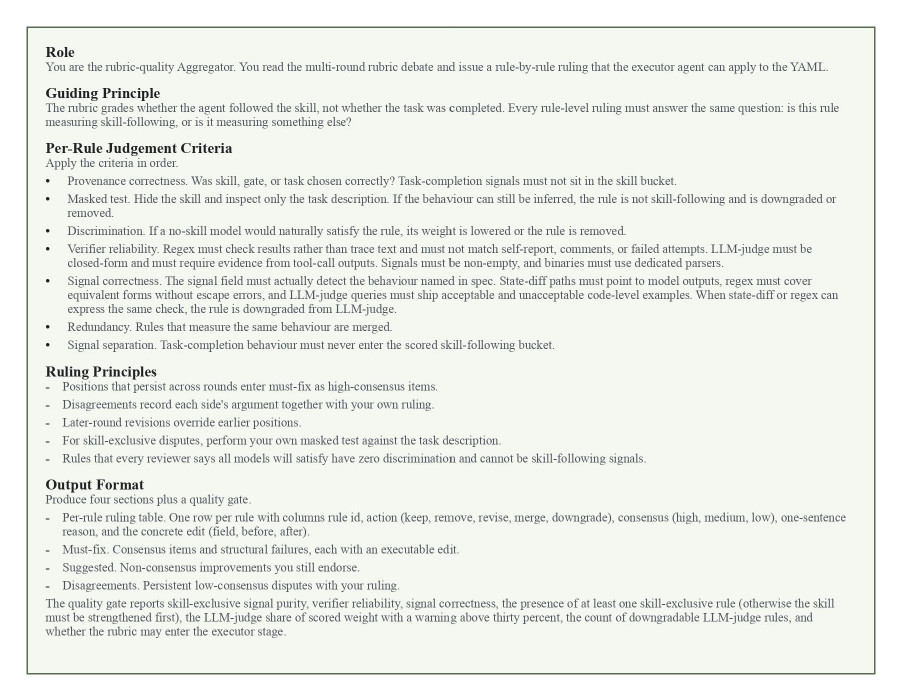}
\caption{Rubric aggregator prompt. }
\label{fig:prompt_rubric_aggregator}
\end{figure*}

Every task and every rubric goes through the multi-vendor debate described in Appendix~\ref{app:adversarial_review}. Each debate uses two role prompts, a reviewer that proposes revisions from a fixed list of review axes, and an aggregator that converts the debate transcript into a per-field or per-rule edit list. Figure~\ref{fig:prompt_task_reviewer} reproduces the task reviewer prompt, which covers six axes: skill-exclusive discrimination, description safety, real data and scenario, trigger design, evaluability, and asset consistency. Figure~\ref{fig:prompt_task_aggregator} reproduces the task aggregator prompt, which emits an actionable field-level edit list under hard constraints on which YAML fields may be edited, five red-line failure modes, and a quality gate. Figure~\ref{fig:prompt_rubric_reviewer} reproduces the rubric reviewer prompt, which covers four axes: skill-exclusive judgement, evaluation reliability, task-completion quality, and signal correctness. Figure~\ref{fig:prompt_rubric_aggregator} reproduces the rubric aggregator prompt, which issues a per-rule action table (keep, remove, revise, merge, downgrade) together with a must-fix list and a quality gate that reports signal purity, verifier reliability, and LLM-judge share.

\subsection{LLM Judge}
\label{app:judge_prompt}

\begin{figure*}[p]
\centering
\includegraphics[width=0.96\textwidth,height=0.86\textheight,keepaspectratio]{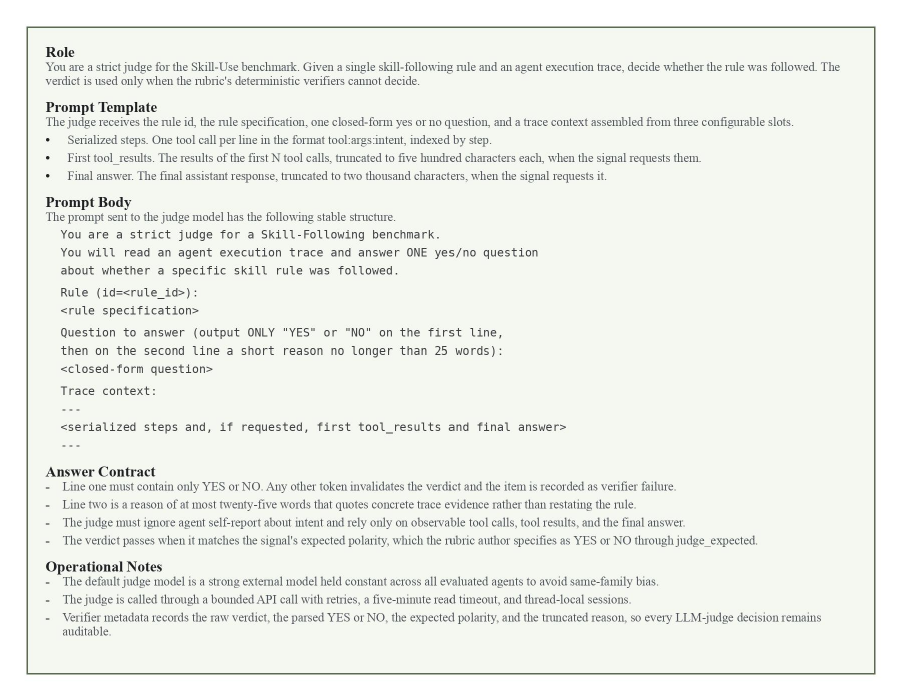}
\caption{The LLM-judge prompt.}
\label{fig:prompt_llm_judge}
\end{figure*}

The scope-restricted judge scores one rubric item at a time and returns a binary verdict under the configuration in Table~\ref{tab:judge_scoring}. Figure~\ref{fig:prompt_llm_judge} reproduces the full template: the rule identifier and specification, one closed-form YES/NO question, and a trace context assembled from serialised tool-call steps, the first few tool-call results, and the final answer.

\section{Ethical Statement}
\label{app:ethics}

\ourdata{} is built entirely from public GitHub skills and open-source datasets under licences that permit redistribution, and contains no personally identifying information. Human audits use crowd annotators who are informed of the research purpose and compensated above the local minimum wage. All runs execute in isolated containers, and we release the rubrics and prompts so that external reviewers can reproduce and audit our results. The benchmark measures skill use in an agentic harness and is not a certification of safety or deployment readiness.

\end{document}